\documentclass[]{interact}
\usepackage{listings}
\usepackage{epstopdf}
\usepackage[caption=false]{subfig}
\usepackage{xcolor} 
\usepackage{booktabs}
\usepackage{tabularx}
\usepackage{array}
\newcolumntype{Y}{>{\raggedright\arraybackslash}X}
\usepackage{xcolor}
\usepackage{graphicx} 
\usepackage{adjustbox}   
\usepackage{url}
\usepackage{longtable}
\usepackage{colortbl}
\makeatletter
\def\LT@makecaption#1#2#3{\LT@mcol\LT@cols c{\hbox to\z@{\hss\parbox[t]\LTcapwidth{%
  \tablecaptionfont\noindent{\tablenumfont#1{#2}}\hskip7pt\ignorespaces#3\par
  \vskip\belowcaptionskip\vskip4pt}\hss}}}
\makeatother

\definecolor{wfA}{HTML}{2E7D5B}   
\definecolor{wfB}{HTML}{2F5F98}   
\definecolor{wfC}{HTML}{B5652B}   
\definecolor{wfMCP}{HTML}{9C7A12} 

\usepackage[longnamesfirst,sort]{natbib}
\bibpunct[, ]{(}{)}{;}{a}{,}{,}
\renewcommand\bibfont{\fontsize{10}{12}\selectfont}

\definecolor{pmBlue}{HTML}{1F4E79}
\definecolor{pmOrange}{HTML}{B85C1E}
\definecolor{pmGreen}{HTML}{2E6B4F}
\definecolor{pmGrey}{HTML}{5A5A5A}
\definecolor{pmGreyLight}{HTML}{EFEFEF}

\lstdefinestyle{promptlst}{
  basicstyle=\ttfamily\scriptsize,
  columns=fixed, basewidth=0.52em, keepspaces=true,
  keywordstyle=\color{pmBlue},
  stringstyle=\color{pmOrange},
  numbers=left, numberstyle=\tiny\color{pmGrey}, numbersep=6pt,
  showstringspaces=false, breaklines=false,
  frame=single, rulecolor=\color{pmGrey!40}, backgroundcolor=\color{pmGreyLight},
  framesep=3pt, xleftmargin=14pt, aboveskip=4pt, belowskip=4pt,
  escapeinside={(*@}{@*)}
}
\lstdefinelanguage{promptMD}{
  alsoletter={_},
  morekeywords={DB_FILENAME, DB_PATH_HINT, Engine, CREATE, TABLE, SELECT, FROM, ORDER, BY, PRIMARY, KEY},
  morestring=[b]",
  moredelim=[l][\bfseries\color{pmBlue}]{\#},
  sensitive=true
}

\theoremstyle{remark}

\begin{document}

\articletype{Research Article}

\title{Agentic AI-Assisted Modeling for Production Scheduling: Assessment in Constraint Programming}

\author{
\name{Ángel Sánchez-Fernández\textsuperscript{a*}\thanks{* Ángel Sánchez-Fernández. Email: angel.sanchez1@udc.es. ORCID: 0009-0008-7139-4736}, Javier Pernas-Álvarez\textsuperscript{a} ORCID: 0000-0001-9185-3583 and Diego Crespo-Pereira\textsuperscript{a} ORCID: 0000-0002-0797-249X}
\affil{\textsuperscript{a}Universidade da Coruña, Campus Industrial de Ferrol, CITENI, Grupo Integrado de Ingeniería, Esteiro s/n, Ferrol, 15403.}
}

\maketitle

\begin{abstract}

Developing optimization models for production scheduling requires substantial expert effort. Research on large language models (LLMs) has followed two directions: specialized approaches for automated modeling, mostly for mixed-integer linear programming, which often rely on dedicated training or problem-specific architectures that limit industrial deployment; and agentic artificial intelligence for operational decision support, which generally assumes that the optimization model already exists. This study bridges both directions by assessing whether general-purpose LLMs, orchestrated as agents without task-specific training, can formulate and implement constraint programming models from natural-language problem descriptions. Single-agent and multi-agent architectures are integrated with a Model Context Protocol server that provides context-aware retrieval of solver documentation to mitigate hallucinations during implementation. Both are compared with a direct LLM baseline on six industry-oriented problems covering flow-shop, job-shop, flexible job-shop and resource-constrained warehouse scheduling, using three LLMs and assessing modeling accuracy, execution success, latency and token consumption. Formulation proves largely within reach of current LLMs, whereas implementation is the main barrier. The multi-agent workflow raises the share of scripts that run correctly as generated from 14.8\% with a direct LLM call to 59.3\%, reaching 80.6\% on the four less complex problems, while tightly coupled intralogistics models remain an open challenge.

\end{abstract}


\section{Introduction}

In recent years, society has undergone a profound transformation driven by the rapid adoption of artificial intelligence (AI) based solutions. At the same time, the benefits of these technologies have become commonplace in academic research and discussions among industry experts in nearly every sector, with Operations Research (OR) emerging as a particularly promising application domain for Large Language Models (LLMs). Despite the maturity of simulation and optimization techniques, a significant gap remains between industrial practitioners and OR experts, since developing optimization models demands specialized expertise in mathematical formulation, problem understanding, requirements definition, coding, and data management, all of which make the modeling process time-consuming, costly, and often inaccessible to non-experts. This challenge has led to the concept of optimization democratization, in which AI serves as a bridge between both stakeholders by reducing technical barriers and enabling non-experts to develop and use optimization models more effectively \citep{Ivanov2026, SimchiLevi2025}. 

A growing stream of recent OR are centered on the emerging concept of Agentic AI, which leverages LLMs as cognitive engines capable of integrating task planning, tool invocation, and reflective reasoning \citep{Jannelli2025, Rossi2026}. Through these capabilities, agentic systems can iteratively decompose and execute complex goals without relying on fully predefined workflows. In the OR domain, most current applications are oriented toward intelligent manufacturing and supply chain management, where AI agents can access relevant system information and augment the capabilities of digital twins and previously formulated mathematical models \citep{Chen2026, Schmitt2026, Ye2026}. Building on these resources, agentic systems can translate operational requirements into optimization scenarios, execute optimization solvers, extract and configure model parameters, and interpret results through natural language. This type of agentic integration closely aligns with the Industry 5.0 paradigm, where autonomy is oriented toward supporting human-centric decision making, human-AI collaboration, and intuitive interaction rather than replacing human responsibility \citep{Ivanov2026, Leng2026, Ye2026}.

Nevertheless, most functionalities of current agent-based solutions still rely on optimization models that have already been developed. This raises several open research questions concerning the distribution of tasks between LLMs and human experts, the extent to which modeling activities can be delegated to autonomous agents, and the mechanisms through which human users can validate, supervise, and effectively exploit their outputs. In this context, a particularly relevant question is whether LLM-based frameworks can construct the optimization model itself from scratch and support the complete modeling workflow, from mathematical formulation to results reporting, including model coding, execution, and interpretation, based solely on problem data and a natural-language problem description. Several recent studies have begun to explore this direction, from which the following research gaps (RG) can be identified:

\begin{itemize}
    \item RG1 - Knowledge engineering bottlenecks: Many high-performing framework architectures, such as SMILO, SyntAGM or OptiMind,  require domain experts to construct the inputs and modeling knowledge used to guide the LLM. This dependence on expert-defined knowledge limits their scalability and transferability across different application domains \citep{LiQ2026, Rossi2026, ref01_Zhang_OptiMind}.
    \item RG2 - Limited Benchmarking: Current evaluation is often hindered by benchmarks that contain ambiguous problem data, incorrect reference answers, or missing parameters. In addition, many benchmarks reported in the literature remain relatively general, relying on small-scale instances and conventional OR problems that have already been extensively studied. Therefore, new benchmarking efforts are needed to assess LLM-based frameworks on more complex, realistic, and demanding optimization problems.
    \item RG3 - Token Consumption and latency: The execution of high-dependency, multi-agent systems, such as Chain-of-Experts architectures, and complex reasoning-based LLM workflows is often constrained by token limits and finite context windows \citep{Rossi2026}. Moreover, these frameworks typically involve substantial token consumption and high latency, which are frequently overlooked despite their direct impact on practical applicability. Therefore, token efficiency should be explicitly considered when developing LLM-based solutions and discussed in relation to the requirements of the target application. From the authors' perspective, this performance indicators are still insufficiently incorporated into robust benchmarking metrics.
    \item RG4 - Hallucination: Since LLMs are probabilistic next-token predictors, they may hallucinate across multiple stages of the optimization process. During mathematical formulation, this can lead to the generation of nonexistent constraints, decision variables, or incorrectly specified objective functions. Similarly, during coding tasks involving specific optimization programming APIs, hallucinations may result in syntactically incorrect code, physically impossible models, or mathematically infeasible formulations.
    \item RG5 - Limited exploration of LLM-optimization interactions: In the context of LLM-assisted frameworks for optimization, significant progress has been made in certain optimization methods, particularly mixed-integer linear programming (MILP). However, the use of LLMs to support the formulation, coding, and execution of CP models remains underexplored. This represents a relevant research gap, as CP has emerged as a strong alternative to MILP and has proven particularly effective for solving large-scale combinatorial problems in industrial settings.
     
\end{itemize}

This work addresses the identified research gaps by advancing the development of AI-assisted frameworks for solving OR problems through human-machine interaction based on natural language. The proposed approach aims to assess how autonomous agents can perform formulation and modeling tasks by delegating complex reasoning and coding to LLMs, while supporting human decision-making through more accessible optimization modeling. To this end, we propose several agentic approaches: First, a multi-agent framework is structured around two key stages of the optimization modeling process: (i) natural-language information processing and model formulation, where the problem description is translated into a set of decision variables, constraints, and an objective function; and (ii) model encoding according to the solver specifications, where the agent can retrieve and validate the required functions, handle parameter formats, and assemble the resulting model in accordance with the formulated problem. Second, a single-agent architecture is developed to assess whether assigning both stages to a single agent provides any benefits compared with the multi-agent approach. Nevertheless, in both approaches, the encoding architecture incorporates an MCP server hosting the solver API documentation. Through a context-retrieval toolset, the agent can identify and retrieve the relevant functions and specifications required for model encoding, with the aim of reducing hallucinations during this second stage.

The experiments conducted in this work aim to assess the extent to which these agentic approaches improve model generation compared with direct LLM-based modeling, while broadening the empirical evaluation of the field through six industry-based problems of varying complexity and a set of relevant performance indicators.

Specifically, this work makes the following contributions in response to the identified research gaps:

\begin{itemize}
\item Scalable multi-agent modeling framework: An AI-agentic framework has been developed for the formulation and modeling of mathematical optimization problems using CP. The proposed framework reduces the degree of expert-in-the-loop involvement through a structured workflow composed of three main stages: preprocessing of the problem description and input data, mathematical model formulation, and encoding for direct implementation. The framework is organized into clearly differentiated modules, enabling its scalability to other optimization techniques (RG1 and RG5).
\item MCP-based tool orchestration for CP: We developed an MCP server with a modular architecture in which the tools exposed to the LLM correspond to predefined methods of the CP API. This enables the LLM to translate the mathematical formulation into a CP model through structured tool selection and invocation. The toolset developed in this work allows the agent to search relevant documentation and modeling examples, and to identify and validate the functions required for model implementation without loading the entire API documentation into the context. This reduces unnecessary contextual information and token usage while helping to mitigate hallucinations during the model encoding and execution stages (RG4 and RG5).
\item LLMs benchmarking across industrial scheduling problems: The proposed framework is tested across six different scheduling problems, including flow-shop, job-shop, flexible job-shop, and three variants of resource-constrained scheduling involving transport and picking tasks in automated warehouses. These problems were selected to represent different problem structures and levels of complexity in terms of the number and type of decision variables, constraints, resources, and operational parameters, thereby challenging the evaluated LLMs under diverse and demanding conditions. The objective is to assess their performance in terms of effectiveness, measured through modeling accuracy, and efficiency, measured through token usage and latency, for both formulation and model execution tasks (RG2 and RG3).

\end{itemize}

The remainder of this paper is organized as follows. Section 2 reviews the literature on constraint programming models for production scheduling, LLM-based approaches to mathematical optimization, and agentic AI applications in scheduling and manufacturing. Section 3 presents the proposed agentic workflows and MCP architecture. Section 4 describes the experimental procedure and presents the experimental results. Section 5 discusses the main findings and their implications. Finally, Section 6 concludes the paper and outlines directions for future research.

\section{Literature Review}

\subsection{Large Language Models for mathematical optimisation}
\label{subsec:LLM for mathopt review}
In recent years, multiple contributions have addressed the generation of mathematical models from natural-language descriptions, with the potential to facilitate the adoption and transfer of mathematical optimization methods to organizations lacking dedicated expertise in this field. In fact, models such as GPT, Claude, DeepSeek, and Gemini have been evaluated on mathematical reasoning benchmarks, showing strong performance in tasks requiring complex reasoning \citep{ref_arxiv2506_00309_Wang2025}. Regarding the use of LLMs and recent advances in the automation of mathematical modeling, different methodological strategies can be distinguished, together with the optimization techniques adopted in each development. In particular, recent contributions on LLM-assisted modeling for mixed-integer linear programming (MILP) problems have demonstrated promising results in the automatic generation of mathematical formulations. 

OptiMind is a Small Language Model (SLM) specifically developed for this purpose through expert-supervised fine-tuning using training data derived from representative OR problem classes. The reported results show that this domain-specific training substantially improves the accuracy of the generated formulations, outperforming its base model, local models such as Qwen3-32B, and even frontier models \citep{ref01_Zhang_OptiMind}. In OptiMUS-0.3, the authors adopt a training-free approach and design a multi-agent workflow compatible with general-purpose LLMs, incorporating retrieval-augmented generation (RAG) capabilities to retrieve domain-specific context and relevant examples that support the formulation process and improve modeling accuracy. Subsequently, SMILO \citep{ref02_Li_SMILO} introduces a graph-based approach for model construction and deterministic formulation assembly, guided by expert-defined modeling knowledge, and substantially outperforms OptiMUS on shift and days-off scheduling problems. However, these developments suggest a clear trade-off between scalability and accuracy when comparing general-purpose solutions for MILP model generation with ad hoc approaches designed for problems of manageable complexity and a specific problem structure. Further relevant studies have proposed integrated frameworks in which agents, RAG mechanisms, LLMs, and domain-specific tools are combined to support the generation of MILP models in industrial settings \citep{ref08_Chen_MultiRobotMILP, ref06_ElBaz_Warehouse, ref05_Jiang_RideAgent, ref07_Hu_PortScheduling}. Regarding the generation of CP models from natural-language descriptions, the literature remains comparatively limited.

Nevertheless, ConstraintLLM \citep{ref07_Shi_ConstraintLLM} has emerged as a fine-tuned model for addressing industrially relevant problems through the PyCSP3 solver, incorporating a tree-of-thought reasoning mechanism and iterative self-correction. The reported results show improved performance compared with general-purpose LLMs such as GPT-4o and DeepSeek-V3. Regarding its broader industrial applicability and scalability, some limitations persist, including its reliance on a specific solver and a training strategy restricted to the Qwen2.5 model family. Other approaches rely on domain-specific languages (DSLs) to constrain LLM outputs and ensure that the generated models are fully executable by the target solver. However, this comes with inherent limitations, particularly due to the need to define problem- and solver-specific semantic rules \citep{ref10_Shi_ConstraintSpecification}. Table~\ref{tab:hybrid_approaches} summarizes the main contributions to the automatic generation of mathematical optimization models using LLMs. The reviewed studies show that approaches for CP model generation remain comparatively limited with respect to techniques such as MILP. They also reveal a clear tendency to trade modeling generality for accuracy, as architectures become increasingly tailored to specific solvers and problem classes in order to mitigate hallucinations. In this context, agentic AI appears to be a promising alternative due to its flexibility in knowledge retrieval, modular architectural design, and iterative reasoning capabilities.

\begin{table*}[htbp]
\centering
\caption{Recent contributions for automated mathematical modelling with LLMs.}
\label{tab:hybrid_approaches}
\resizebox{\textwidth}{!}{
\begin{tabular}{llccc}
\toprule
\textbf{Work} & \textbf{Approach Summary} & \textbf{Method} & \textbf{Train} & \textbf{Scalability (Sol $\mid$ Prob $\mid$ LLM)} \\
\midrule
OptiMind & SLM fine-tuned and hint injection & MILP & $\checkmark$ & $\ominus$ $\mid$ $\bullet$ $\mid$ $\circ$ \\
OptiMUS-0.3 & Modular agent with constraint graph \& reflective correction & MILP & $\times$ & $\bullet$ $\mid$ $\bullet$ $\mid$ $\bullet$ \\
SMILO & Modelling graph-guided tasks \& template assembly & MILP & $\times$ & $\ominus$ $\mid$ $\ominus$ $\mid$ $\bullet$ \\
Chen et al. & Domain specialization via knowledge-augmented fine-tuning & MILP & $\checkmark$ & $\ominus$ $\mid$ $\circ$ $\mid$ $\circ$ \\
El Baz et al. & Agentic model with human-in-the-loop OR expert & MILP & $\times$ & $\ominus$ $\mid$ $\circ$ $\mid$ $\bullet$ \\
RideAgent & Small-sample guided heuristic agent for variable fixing & MILP & $\times$ & $\ominus$ $\mid$ $\circ$ $\mid$ $\bullet$ \\
Hu and Li & GAR/GPVSS with Feasibility-Aware Constrained Decoding (FACD) & MILP & $\checkmark$ & $\circ$ $\mid$ $\circ$ $\mid$ $\bullet$ \\
ConstraintLLM & LLM fine-tuned with Constraint-Aware Retrieval Module & CP & $\checkmark$ & $\ominus$ $\mid$ $\bullet$ $\mid$ $\circ$ \\
Shi et al. & Domain-Specific Language with dual program views for JSP & CP/MILP & $\times$ & $\bullet$ $\mid$ $\bullet$ $\mid$ $\circ$ \\
\bottomrule
\multicolumn{5}{p{\textwidth}}{\footnotesize \textit{Legend:} \textbf{Train}: $\checkmark$ = Uses base LLM training/SFT; $\times$ = No training required. \textbf{Scalability}: \textbf{Sol} = Solver, \textbf{Prob} = Problem class, \textbf{LLM} = LLM generalization. \textbf{Symbols}: $\bullet$ = Highly scalable (Directly adaptable); $\ominus$ = Medium (Reasonable work needed); $\circ$ = Low scalability (Completely ad-hoc/difficult to scale).} \\
\end{tabular}
}
\end{table*}

\subsection{Constraint Programming for Production Scheduling}

Given the substantial body of recent work focused on MILP-based approaches, it is also relevant to explore alternative optimization paradigms such as CP, which has demonstrated strong applicability to NP-hard and combinatorial optimization problems arising across a wide range of scheduling domains. In many of these settings, CP has shown competitive performance with respect to mixed-integer programming (MIP) and MILP approaches, particularly for highly constrained and large-scale problem instances.

Existing contributions can be grouped according to several problem classes of industrial relevance, for which CP-based procedures have been evaluated on instances whose scale considerably exceeds that of traditional OR formulations and benchmark settings. In the context of the job-shop scheduling problem (JSSP), common CP formulations rely on arrays of interval variables combined with precedence and non-overlap constraints. Both commercial and open-source CP engines have been evaluated, including IBM ILOG CP Optimizer (CPO) {\citep{Laborie2018CPO}} and the CP-SAT solver from OR-Tools {\citep{Perron2023CPSAT}, together with hybrid platforms such as Hexaly, formerly LocalSolver, which combines local search with constraint propagation and linear relaxations behind a single modelling interface \citep{Benoist2011LocalSolver, Hexaly2026}}. Reported results highlight the scalability of these solvers, with CPO being able to address instances involving up to approximately one million operations, thereby illustrating the potential of CP for large-scale industrial applications such as semiconductor manufacturing \citep{ref13_DaCol_IndustrialJobShop}. A recent controlled benchmark of CPO, CP-SAT, Hexaly and OptalCP on 332 JSSP instances of up to $1000 \times 1000$ operations confirms this scalability and shows that no single engine dominates: OptalCP and CPO certify the largest number of optima, whereas Hexaly returns the best schedules on industrial-scale instances with tens of thousands of operations, which supports an instance-aware choice of solver \citep{Yuraszeck2026CPBenchmark}, an idea also exploited through machine-learning-based solver selection for the flexible job shop \citep{Mueller2022SolverSelection}. When the same job-shop models are formulated for MILP and CP solvers, CP consistently attains the smallest deviations from the best-known solutions on medium and large instances, whereas MILP solvers remain competitive only on small ones \citep{Naderi2023MIPvsCP}. Within the JSSP family, other variants also incorporate transportation between workstations by explicitly integrating task allocation to multiple robots. For benchmark instances from the literature, CP-based formulations have been reported to achieve strong computational performance, outperforming previous MIP formulations \citep{ref18_Ham_TransferRobot}. On the other hand, relevant applications can also be found within the Flexible Job Shop Scheduling Problem (FJSP) family. One study addresses an FJSP variant with compatible job families, in which the number of jobs that can be processed simultaneously is constrained by equipment capacity \citep{ref12_Ham_FlexibleJobShop}. In this case, MIP formulations were compared with CPO on medium- and large-scale instances, with CPO showing clearly superior computational performance and reaching optimal solutions within relatively short solution times. This problem class has also been investigated in distributed manufacturing networks, where CP-based approaches have been reported to outperform four different MILP formulations and metaheuristic methods \citep{ref16_Meng_DistributedFlexibleJobShop}. Hybrid platforms have likewise been adopted where exact MILP models cease to be viable: in the lot-sizing and scheduling of metal parts for car seats, a Hexaly model complements a MIP formulation and GRASP and VNS metaheuristics to minimise changeovers on the largest instances \citep{Colmenar2024Changeover}.

Beyond manufacturing, CP has also demonstrated relevant applications in warehouse scheduling and allocation problems arising in intralogistics optimization, where operational decisions must be coordinated under strict time-window constraints. The literature includes CP formulations incorporating integer assignment variables that outperform previous MIP formulations and achieve optimal solutions within fractions of a second \citep{ref15_Telha_StorageSpace}. Within the same application domain, CP has also been integrated with complementary optimization tools to jointly address order-line batching, batch scheduling under deadlines, and picker routing within a unified framework \citep{ref17_Haouassi_OrderlineBatching}. This integration of multiple optimization tools with CP for highly demanding real-world problems has been successfully applied in several settings. In the context of FJSP in the shipbuilding industry, a two-stage CP-based decomposition has been proposed. In the first stage, the temporal scheduling of operations is addressed according to due-date requirements while minimizing the makespan. In the second stage, the spatial allocation of resources and working areas is optimized, with the proposed approach outperforming corresponding MILP formulations \citep{ref19_PernasAlvarez_Shipbuilding}. Decomposition strategies have also been proposed for the JSSP, including approaches in which an initial phase generates a warm start for a subsequent stage, where relaxations are introduced on temporal constraints \citep{ref14_Yuraszeck_ShopScheduling}.

Overall, the reviewed literature highlights the strong capability of CP-based solvers to address NP-hard problems, together with their flexibility in model formulation and their suitability for decomposition-based solution strategies. These characteristics make CP particularly attractive for computationally demanding industrial environments.

\subsection{Agentic AI in scheduling}

Within this context, it is useful to distinguish the main roles currently assigned to agentic AI in industrial scheduling applications. On the one hand, some workflows employ agents as active generators of optimization models, where they contribute directly to problem formulation, model construction, and solver integration. On the other hand, several approaches assume that the underlying optimization or simulation model has already been developed. In these cases, agents operate around the existing model by supporting tasks such as data interpretation and contextualization, experiment execution, result analysis, or decision support at the operational level. This distinction helps clarify whether agentic AI is being used to automate the modeling process itself or to enhance the interaction with and exploitation of previously established optimization and simulation models.

Regarding the first category, apart from the LLM-based approaches discussed in Section~\ref{subsec:LLM for mathopt review}, most of which are not agentic, only one of the reviewed studies focuses on an agent capable of generating MILP models in Python from natural-language specifications, incorporating a human-in-the-loop mechanism between the optimization and simulation stages \citep{ref06_ElBaz_Warehouse}. Beyond this contribution, another agentic approach has been proposed for constraint programming, in which an MCP-based architecture connects the LLM with symbolic solvers to enable autonomous CP model generation, reporting 100\% accuracy across 101 benchmark problems \citep{ref20_Szeider_CPAgent}. Despite the clear advantages of these agentic approaches for automated model generation, their validation remains largely limited to benchmark settings, with limited evidence of industrial applicability. In contrast, multi-agent frameworks have also been evaluated directly in smart manufacturing environments by implementing sequential workflows and combining RAG mechanisms with chain-of-thought reasoning to retrieve relevant knowledge from software libraries and support model generation under dynamic production requirements \citep{ref11_Li_A4PS}. 

Regarding the second category, a broader range of agentic AI applications and use cases can be identified \citep{ref20_Chen_LLM_SmartManufacturing}. In these approaches, agents are typically employed either to contextualize operational data and physical system requirements in order to support decision-making and trigger actions on pre-existing models \citep{Ye2026, ref22_Xiang_WaferScheduling}, or as conversational assistants capable of connecting to databases and internal tools to provide decision support at the shop-floor level \citep{ref24_Jannelli_AgenticSupplyChain, ref23_Yuan_ChatMES}.

Taken together, the reviewed literature shows that LLM-based agentic tools have been applied mainly to operational and shop-floor tasks, while their use for mathematical formulation and optimization modeling remains largely unexplored. This gap motivates the investigation of modular agentic architectures that can be adapted to different optimization paradigms and solvers while streamlining the modeling process.

\section{Methodology}
The methodology adopted in this work is structured around two main sections: Section~\ref{subsec:agentic-framework} presents the evaluated agentic framework, while Section~\ref{subsec:MCP-architecture} describes the MCP-based tool architecture.

\subsection{Agentic Framework}
\label{subsec:agentic-framework}

\begin{figure}[!h]
\centering
\includegraphics[width=0.95\linewidth]{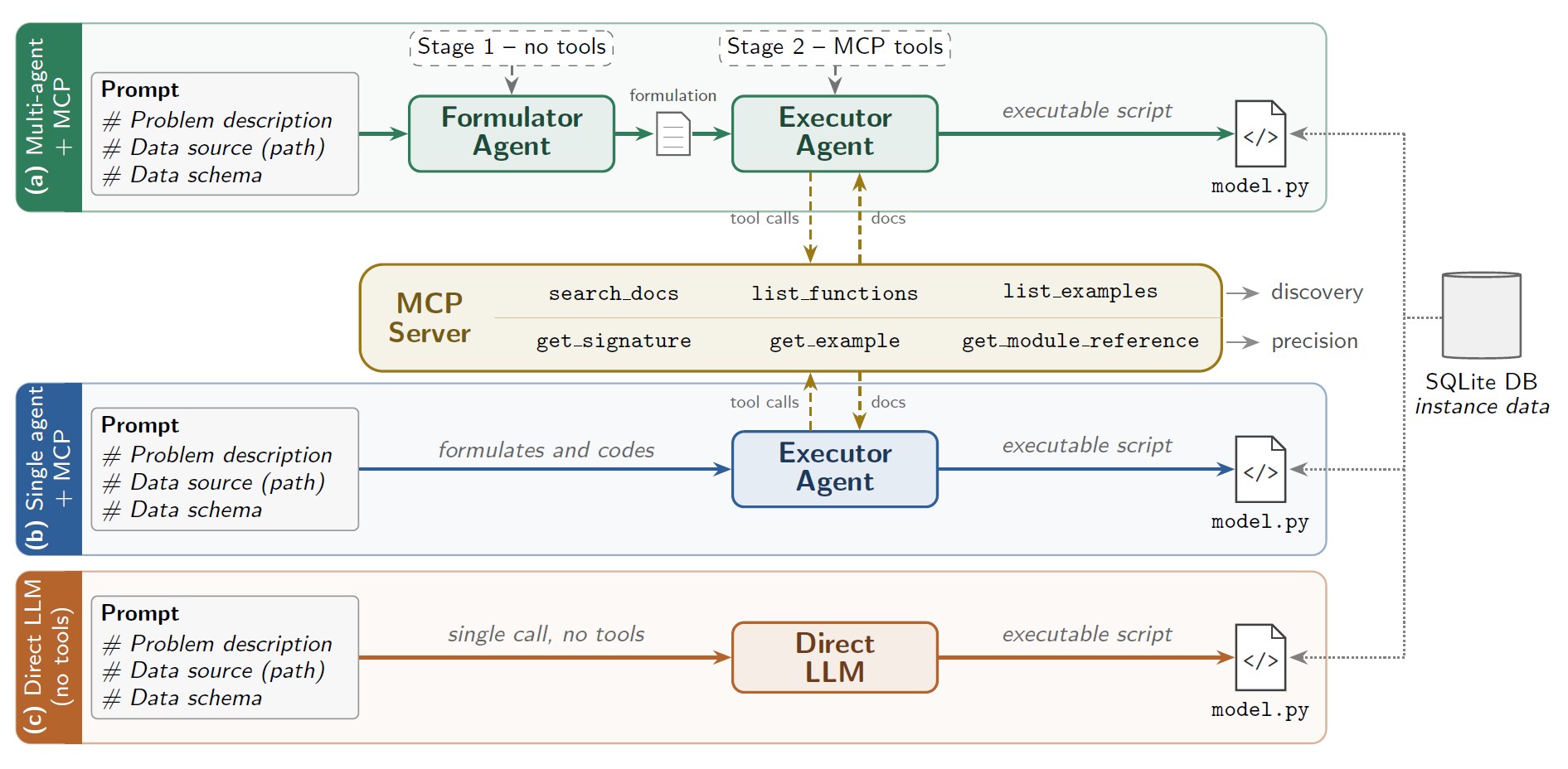}
\caption{Three workflows compared in this study. (a) Two-stage multi-agent pipeline: a tool-free Formulator Agent hands a formulation to an Executor Agent connected to the MCP documentation server. (b) Single Executor Agent connected to the same MCP server. (c) Direct LLM baseline without tool access.}
\label{fig:workflows}
\end{figure}

The proposed framework is implemented in three workflows (Figure~\ref{fig:workflows}) that differ only in how the modelling task is distributed: (a) a multi-agent pipeline in which formulation and coding are assigned to two specialised agents, (b) a single agent that performs both tasks, and (c) a baseline in which the LLM completes the whole task in a single call without tool access. Workflows (a) and (b) are connected to an MCP server that exposes the documentation of \texttt{docplex.cp}, the Python modelling interface of IBM ILOG CP Optimizer. All workflows receive the same problem prompt and return a standalone Python script that reads the instance data from an SQLite database, builds and solves the CP model, and reports the solution.

The workflows were built in Langflow Desktop (version 1.11.6), and was chosen because it is built on LangChain and LangGraph, the de facto open-source for constructing LLM agents, and thus keeps the workflows independent of any single provider. Table~\ref{tab:agent-config} summarises the components of each workflow and their key settings. To isolate the effect of the architecture, the sampling parameters are identical in all workflows, the same LLM is assigned to every component within a run, and each component uses the same system prompt for every LLM. All prompts, output contracts and instance data are publicly available in the project repository~\citep{AgenticPScheduling2026}.

\begin{table}[!h]
\centering
\caption{Langflow components of each workflow. All components share the same LLM settings: temperature 0.15 and a limit of 15\,000 output tokens per call.}
\label{tab:agent-config}
\footnotesize
\begin{tabularx}{\linewidth}{@{}l l Y c c@{}}
\toprule
Workflow & Component & Task & MCP tools & Max.\ iterations$^{\rm a}$ \\
\midrule
\textcolor{wfA}{\textbf{(a) Multi-agent}}  & Formulator Agent & Formulation and implementation plan & --  & 10 \\
                                           & Executor Agent   & Coding of the given formulation     & 6   & 50 \\
\addlinespace
\textcolor{wfB}{\textbf{(b) Single agent}} & Executor Agent   & Formulation and coding              & 6   & 50 \\
\addlinespace
\textcolor{wfC}{\textbf{(c) Direct LLM}}   & LLM call         & Formulation and coding              & --  & 1  \\
\bottomrule
\end{tabularx}
\begin{tabnote}
$^{\rm a}$\,Maximum number of LLM calls in the agent loop; each call either requests tool calls or returns the final answer.
\end{tabnote}
\end{table}

The components of the framework are described below.

\begin{figure}[!htbp]
\centering
\includegraphics[width=0.95\linewidth]{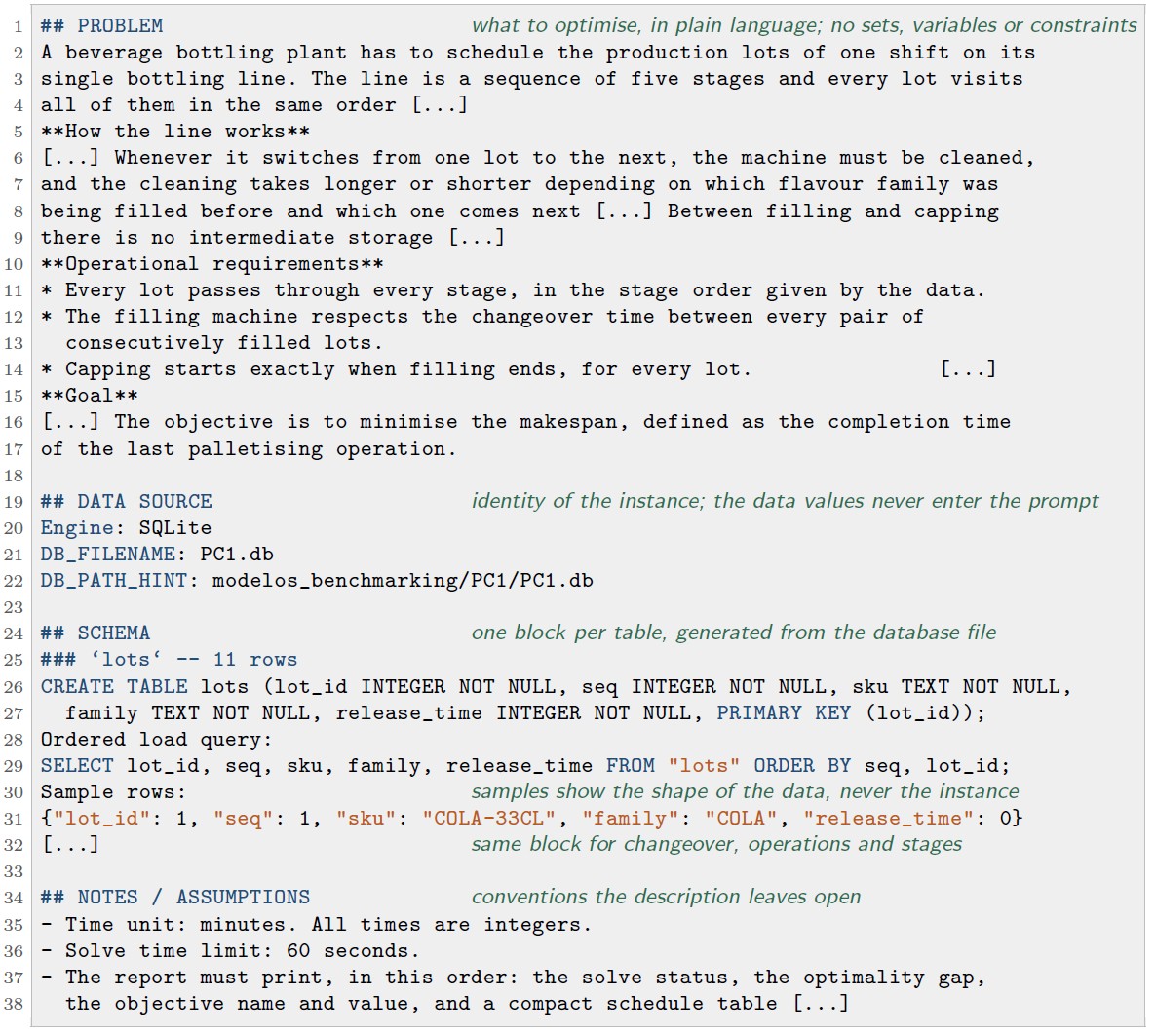}
\caption{Structure of the prompts. The task message that enters every workflow, shown for problem PC1: four sections in a fixed order, with the process described in natural language only, the identity of the SQLite instance, the schema of every table generated from the database file, and the conventions left open by the description. The complete prompts and the task messages of PC1--PC6 are provided as supplementary material and in the project repository \citep{AgenticPScheduling2026}.}
\label{fig:prompt-structure}
\end{figure}

\begin{itemize}
    \item Prompt: Figure~\ref{fig:prompt-structure} shows how the prompts are designed. The input to every workflow is a task message with four sections, identical for all workflows and LLMs. The problem description states in natural language the process to be scheduled, its operational requirements and its objective, without any formulation guidance: sets, decision variables, constraints and the type of model must be identified entirely by the LLM. The data source gives the file name and path of the SQLite database that holds the instance, and the data schema describes each of its tables with the exact \texttt{CREATE TABLE} statement, the row count, an ordered load query and three sample rows, all generated from the database file; the instance values themselves are never placed in the prompt, so the LLM must write the code that retrieves them at execution time, which keeps the context small and independent of the instance size. The system prompts used by the agents (workflows (a) and (b) in Figure~\ref{fig:workflows}) and by the single LLM (workflow (c)) are available in the repository \citep{ref_repo_AgenticPScheduling}. These system prompts are designed to define the role of each component.

    \item Formulator Agent: Used only in workflow (a), this agent translates the problem description into a structured specification for the Executor Agent. It has no tool access and works under a system prompt that defines its role, its input and a three-part output contract: (i) the mathematical formulation, with sets, parameters, decision variables, labeled constraints and objective; (ii) a data contract with the SQL queries needed to load each parameter; and (iii) an execution plan for implementing the model.

    \item Executor Agent: This agent produces the final script. In workflow (a) it implements the formulation received from the Formulator Agent, whereas in workflow (b) it also formulates the problem. Its system prompt defines the input, the available MCP tools, a four-step protocol for their use and the structure of the script, which consists of three blocks: data acquisition, CP model, and results report. Before writing any code, the agent must use the tools to verify every \texttt{docplex.cp} function it intends to call. Owing to the tool calls and the documentation added to its context, this is the most expensive component of the framework.

    \item Direct LLM: In workflow (c), in addition to the user prompt, the LLM receives a system prompt that asks for the formulation and the script, with the same three-block structure, in a single response.

    \item MCP Server: The server gives the agents in workflows (a) and (b) access to the \texttt{docplex.cp} documentation through six tools organised in two levels. Three \textit{discovery} tools locate relevant content: keyword search, the symbol index of a module, and a catalogue of solved modelling examples. Three \textit{precision} tools then return the full reference of the selected items: function signatures and parameters, complete example models, and module references. Checking the API before coding reduces hallucinated function calls, at the cost of additional tokens. The server is described in Section~\ref{subsec:MCP-architecture}.

    \item SQLite database: Each of the six problems is stored in its own SQLite database, whose schema matches the one described in the prompt. The generated script is required to open the database in read-only mode and to load all instance data through SQL queries, with no hard-coded values.
\end{itemize}

\subsection{MCP architecture}
\label{subsec:MCP-architecture}

The Executor Agent is designed to generate a script that runs successfully on the first attempt, since the objective of this work is to assess whether agentic approaches can directly produce a valid optimization model without relying on iterative correction procedures. Two main failure modes are considered. The first is formulation error, which is particularly critical because an incorrect mathematical representation of the problem may lead to invalid or misleading decision-support outcomes. The second concerns hallucinations in the use of the solver API, including plausible but non-existent functions, incorrectly named or ordered arguments, and constructs borrowed from related libraries such as \texttt{docplex.cp}. These errors were frequently observed during the preliminary experimentation. The documentation server described in this subsection is, therefore, designed to make verification of the \texttt{docplex.cp} API more accessible and less costly for the agent than relying on unsupported assumptions. Table~\ref{tab:mcp-terms} defines the terms used in the rest of this subsection.

\begin{table}[!htb]
\centering
\caption{Nomenclature of the documentation corpus, illustrated with the method \texttt{solve} of class \texttt{CpoModel}.}
\label{tab:mcp-terms}
\footnotesize
\renewcommand{\arraystretch}{1.15}
\begin{tabularx}{\linewidth}{@{}l Y >{\raggedright\arraybackslash}p{5.3cm}@{}}
\toprule
Term & Definition & Example \\
\midrule
Module         & A \texttt{docplex.cp} sub-package, documented in one reference file (four in the corpus) & \texttt{docplex.cp.model} \\
Symbol         & A documented function, class or method (186, 15 and 132 in the corpus) & method \texttt{solve} of class \texttt{CpoModel} \\
Qualified name & Unique path of a symbol: module, class (if any) and name. Short names may be ambiguous & \texttt{docplex.cp.model.CpoModel.solve}\newline short name \texttt{solve}; \texttt{add} matches two classes \\
Chunk          & Retrieval unit: the complete entry of one symbol, or one whole example script, stored under a deterministic identifier & \texttt{api:docplex.cp.model:}\newline\texttt{docplex.cp.model.CpoModel.solve} ($\approx$535 tokens)\newline \texttt{example:flow\_shop\_permutation} \\
\bottomrule
\end{tabularx}
\end{table}

The architecture follows one principle: the agent loads only the documentation it needs, when it needs it. Context windows vary widely across providers, from a few thousand tokens in some locally served models to about one million in the largest commercial ones, and a tool-calling loop re-sends the whole context at every turn. Loading the whole corpus (77k tokens), or even a plain symbol index (9k), would penalise smaller models and inflate every call, although a typical formulation uses about twenty of the 333 documented symbols. Retrieval thus works in two stages: discovery maps modelling concepts to candidate names across all modules, so the agent never guesses where a symbol lives, and precision returns complete entries for the selected symbols. Both stages are built automatically from the documentation, without embeddings or an expert-curated taxonomy, so the same design can be applied to another solver's API by parsing its reference documentation. Retrieval is also deterministic: identical queries always return identical results, so differences between runs reflect the LLM rather than the retriever.

The resulting server comprises of Python in three modules and is built on the FastMCP interface of the official MCP Python SDK (version~1.27.0) \citep{MCPPythonSDK}. It runs locally and is exposed to Langflow through the Server-Sent Events (SSE) transport. The corpus is parsed into memory at the first tool call and cached, and the only state kept is a set of call counters used for logging and steering.

\begin{figure}[!htb]
\centering
\begin{adjustbox}{max width=\linewidth}
\includegraphics{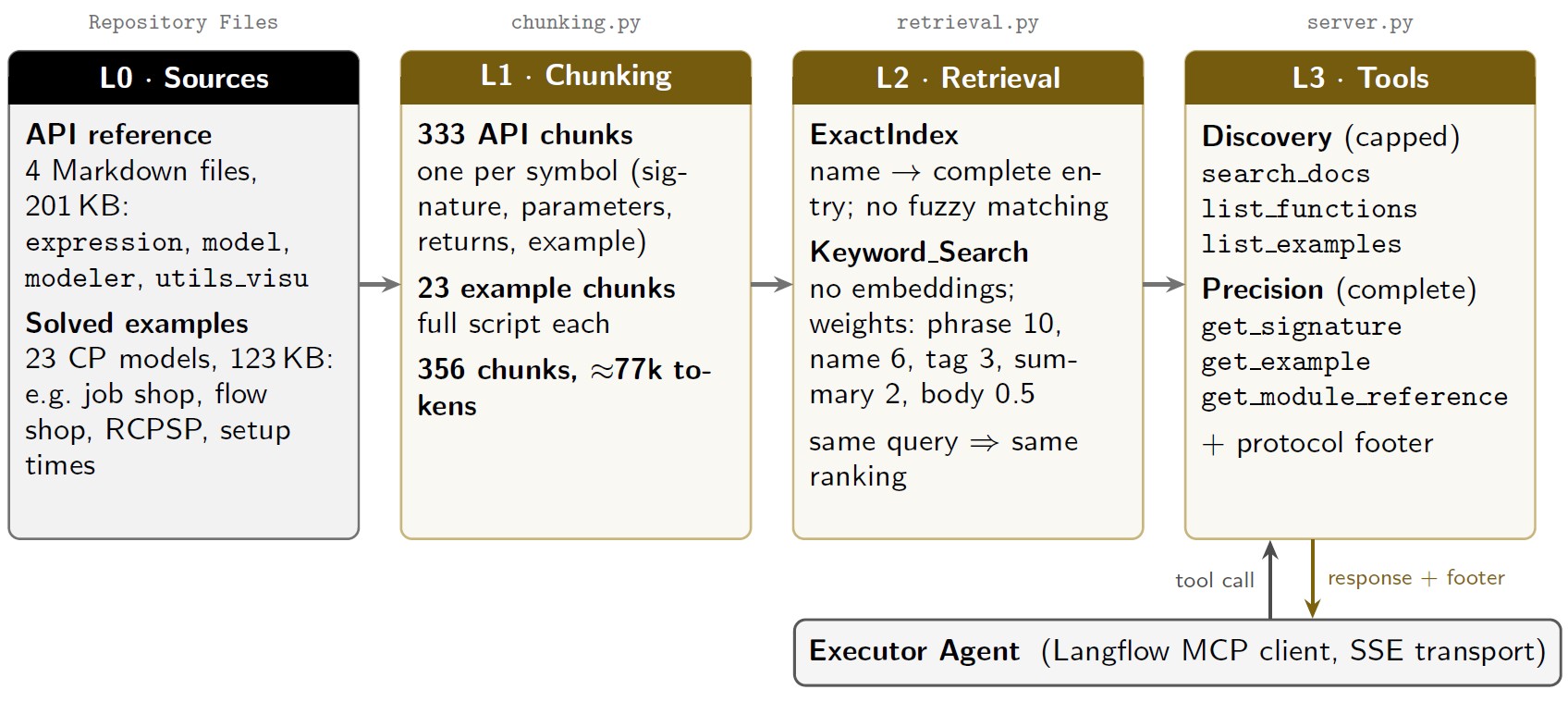}
\end{adjustbox}
\caption{Layered architecture of the documentation MCP server. Token counts are estimated at four characters per token. Every tool response ends with a footer stating the next required step of the retrieval protocol.}
\label{fig:mcp-layers}
\end{figure}

The server is organised in four layers (Figure~\ref{fig:mcp-layers}). 
\begin{itemize}
    \item The source layer (L0) comprises the four Markdown files of the \texttt{docplex.cp} reference, derived from IBM's official documentation, and 23 solved example models, all read in place from the repository.
    \item The chunking layer (L1) applies one structural parser per source type. Reference files are split at symbol headers, so that each of the 333 API chunks keeps a class, function or method together with its signature, parameters, return value and usage example; methods are qualified with their class (e.g.\ \texttt{CpoModel.add}), which is how the agent refers to them. Example scripts are kept whole, because their value lies in the complete model structure.
    \item The retrieval layer (L2) offers two complementary mechanisms: \texttt{ExactIndex} returns the complete entry for a qualified or short name and never approximates, so a name is either found or reported as missing, whereas \texttt{keyword\_search} ranks chunks by weighted token matches, favouring hits on symbol names over tags, summaries and body text.
    \item The tool layer (L3) exposes the six tools described in Table~\ref{tab:mcp-tools}.
\end{itemize}   

\begin{table}[!htb]
\centering
\caption{Tools exposed by the MCP server. Discovery responses are non-authoritative and capped at 400 characters per summary line and 40 candidates per call; precision responses are authoritative and never truncated.}
\label{tab:mcp-tools}
\footnotesize
\renewcommand{\arraystretch}{1.15}
\begin{tabularx}{\linewidth}{@{}l l >{\raggedright\arraybackslash}p{3.3cm} Y@{}}
\toprule
Family & Tool & Input $\rightarrow$ Output$^{\rm a}$ & Key features \\
\midrule
\begin{tabular}[t]{@{}l@{}}\textbf{Discovery}\\ \textit{capped}\end{tabular}
  & \texttt{search\_docs}           & [concepts]~$\rightarrow$ candidate~names     & Merges all queries and groups hits by module; ends with the ready-made \texttt{get\_signature} call \\
  & \texttt{list\_functions}        & module~$\rightarrow$ symbol~index            & Browsing fallback when keyword search finds nothing \\
  & \texttt{list\_examples}         & problem~text~$\rightarrow$ ranked~examples   & Ranked by similarity to the description; alphabetical without one \\
\addlinespace[4pt]
\begin{tabular}[t]{@{}l@{}}\textbf{Precision}\\ \textit{complete}\end{tabular}
  & \texttt{get\_signature}         & [names]~$\rightarrow$ complete~entries       & Order kept; unknown names form a gap list, never approximated \\
  & \texttt{get\_example}           & [ids]~$\rightarrow$ full~source              & Complete working models, up to $\approx$5k tokens each \\
  & \texttt{get\_module\_reference} & module (+\,[names])~$\rightarrow$ reference  & For many symbols of one module; $\approx$28k tokens for \texttt{modeler} \\
\bottomrule
\end{tabularx}
\begin{tabnote}\raggedright
$^{\rm a}$\,Square brackets denote an array argument, so one call resolves several items. The server also accepts JSON arrays inside a string, Python-style lists, comma-separated lists and single names; names are split on commas and spaces, whereas queries keep their internal spaces.
\end{tabnote}
\end{table}

The six tools form two families with explicitly declared authority. Discovery tools map concepts to candidate names; their responses are capped and labeled \texttt{DISCOVERY ONLY}, and the agent is instructed not to write a call from them. Precision tools return complete API entries or example models and are never truncated. Three tools accept arrays, so that every concept or symbol of a formulation can be resolved in a single call. This follows from the cost structure of a tool-calling loop: the whole conversation is re-sent at every turn, so $N$ calls cost roughly $N$ times the base context, however short each response is. Fewer calls therefore save far more tokens than shorter responses.

\begin{figure}[!htb]
\centering
\includegraphics[width=0.95\linewidth]{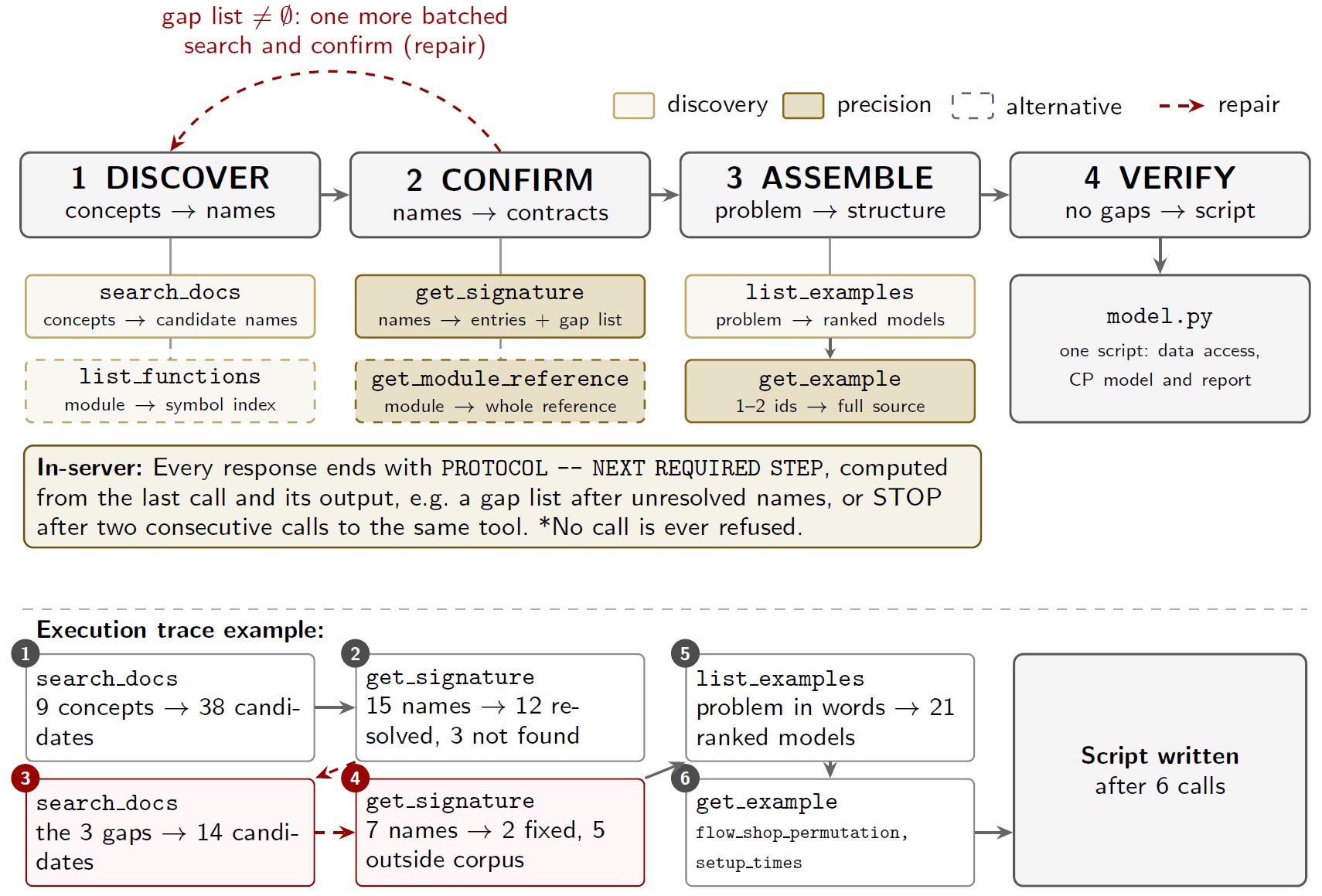}
\caption{Retrieval protocol of the Executor Agent. Top: the four steps, the tools used at each step and the in-server enforcement. Bottom: example of the tool-call trace; calls 3--4 form a repair cycle triggered by the gap list returned in call 2.}
\label{fig:mcp-protocol}
\end{figure}

The way the agent is expected to combine these tools is fixed by a four-step retrieval protocol, shown in the upper part of Figure~\ref{fig:mcp-protocol}. Before writing any code, the Executor Agent (1) searches all the concepts of the formulation in one \texttt{search\_docs} call, (2) confirms every symbol it intends to use in one \texttt{get\_signature} call, (3) describes its problem to \texttt{list\_examples} and loads the one or two structurally closest models with \texttt{get\_example}, and (4) checks that no symbol remains unresolved before writing the script. Since compliance with system-prompt instructions may vary across LLMs, the protocol is also enforced at the server level. Accordingly, each tool response concludes with a \texttt{PROTOCOL -- NEXT REQUIRED STEP} section, dynamically generated from the executed call and its output. For example, after a \texttt{get\_signature} call returns unresolved function names, the server reports them as a gap list and instructs the agent to resolve them through an additional batched retrieval call.

The lower part of Figure~\ref{fig:mcp-protocol} illustrates the protocol in practice through a specific example executed under workflow (a), showing the sequence of calls issued by the Executor Agent. The agent first translated the mathematical formulation into nine CP concepts and then validated 15 API symbols in a single batched call. Three did not resolve: two because the agent had assigned them to the wrong module, and one because \texttt{docplex.cp.solution} is not part of the corpus. The gap list induced a short repair cycle (calls 3--4) that corrected the module errors and made the remaining gaps explicit instead of leaving them as silent guesses. Finally, the agent ranked the examples with a plain-language description of its problem and loaded the two models that together cover the flow-shop structure and the sequence-dependent setups. Finally, the script was written after six calls.

\section{Experimental evaluation}
\label{sec:results}

\subsection{Design of Experiments}

One of the primary aims of this work is to assess the capability of current LLMs and the previously defined agentic approaches to autonomously generate CP models for real-world industrial scheduling problems. To evaluate and challenge the proposed agentic workflows together with the MCP toolkit, the experimental setup considers the following factors and parameter variations:

\begin{table}[!htb]
\centering
\caption{Full-factorial experimental design and run nomenclature.}
\label{tab:doe}
\footnotesize
\renewcommand{\arraystretch}{1.15}
\begin{tabularx}{\linewidth}{@{}l Y l r@{}}
\toprule
Factor & Levels & Codes$^{\rm a}$ & $n$ \\
\midrule
Workflow      & (a) multi-agent + MCP; (b) single agent + MCP; (c) direct LLM (Figure~\ref{fig:workflows}) & WA, WB, WC & 3 \\
LLM$^{\rm b}$ & Gemini 2.5 Pro; GPT-5; DeepSeek V4 Pro & Gem, GPT, DS & 3 \\
Problem       & PC1--PC6 (Table~\ref{tab:problems}) & PC1, \dots, PC6 & 6 \\
Replicate     & Independent generations & r1, r2, r3 & 3 \\
\midrule
\multicolumn{3}{@{}l}{Runs ($3 \times 3 \times 6 \times 3$)} & 162 \\
\bottomrule
\end{tabularx}
\begin{tabnote}\raggedright
$^{\rm a}$\,Each run is identified as \{workflow\}\_\{LLM\}\_\{problem\}\_\{replicate\}; An asterisk aggregates over a field: WB\_*\_PC3\_* denotes all single-agent runs on PC3.
$^{\rm b}$\,All three LLMs are accessed through the OpenRouter API (\texttt{google/gemini-2.5-pro}, \texttt{openai/gpt-5}, \texttt{deepseek/deepseek-v4-pro}).
\end{tabnote}
\end{table}

The LLMs were selected to cover the model families that recur in the literature reviewed in Section~\ref{subsec:LLM for mathopt review}, including both proprietary and open-weight models. GPT, Gemini and DeepSeek models are the usual reference points in evaluations of mathematical reasoning and optimisation modelling \citep{ref_arxiv2506_00309_Wang2025}, and earlier versions such as GPT-4o and DeepSeek-V3 serve as baselines for CP model generation \citep{ref07_Shi_ConstraintLLM}. The design therefore includes their current-generation successors: two proprietary frontier models, Gemini 2.5 Pro and GPT-5, and DeepSeek V4 Pro (open-weight). All three are general-purpose models that support native tool calling and are used without fine-tuning, which is consistent with the training-free and solver-agnostic scope of the framework.

The six problems (Table~\ref{tab:problems}) span the main structures of production and intralogistics scheduling with increasing modelling difficulty. PC1--PC3 cover the three classical shop configurations, each extended with features that are common in practice but absent from textbook instances. PC1 represents a beverage bottling line as a five-stage flow shop with release dates, sequence-dependent cleaning changeovers on the filling machine and a no-wait coupling between filling and capping. PC2 represents a precision machining workshop as a job shop with individual routings, release dates and due dates, and is the only problem that minimises total weighted tardiness instead of the makespan. PC3 is a flexible job shop in an injection moulding plant: each operation can only run on the presses that hold its mould, processing times depend on the machine, and a shared pool of three operators acts as a cumulative resource across all machines. These three instances are synthetic, generated with fixed seeds and sized so that CP Optimizer proves optimality within the 60~s limit. PC4--PC6 reproduce, with their synthetic data, three increasingly detailed CP models of an automated order-picking cell developed in previous work: a basic configuration with one AGV and one picking station (PC4), a variant with limited pallet stock in which the model also decides the integer quantity of each kit type taken from each pallet (PC5), and an extended cell with two AGVs, a capacitated buffer and two picking robots (PC6). Across the benchmark, the reference model grows from a few dozen variables and constraints in PC1--PC3 to more than 3\,000 of each in PC6, and its variables evolve from mandatory intervals only to optional intervals combined with integer and binary assignment decisions.

Each problem comes with a reference model written by the authors in \texttt{docplex.cp} and solved with CP Optimizer under the same 60~s limit imposed on the generated scripts. The reference objectives of PC1--PC4 are proven optimal. Those of PC5 and PC6 are best-known values that did not improve in longer runs of 250--300~s. The schedules returned by the reference models were also checked against every constraint of their problem independently of the solver, including machine overlaps, no-wait couplings, demand satisfaction and buffer capacity.

\begin{table}[!htb]
\centering
\caption{Benchmark problems: structure, size of the reference CP model and ground-truth result.}
\label{tab:problems}
\scriptsize
\setlength{\tabcolsep}{3pt}
\renewcommand{\arraystretch}{1.12}
\begin{tabularx}{\linewidth}{@{}l Y *{6}{>{\centering\arraybackslash}p{1.45cm}}@{}}
\toprule
 & & PC1 & PC2 & PC3 & PC4 & PC5 & PC6 \\
\midrule
\textbf{Problem} & Type$^{\rm a}$ & FSP & JSP & FJSP & RCS & RCS & RCS \\
\addlinespace[3pt]
\textbf{Sets} & Jobs / orders & 11 lots & 10 orders & 6 batches & 29 orders & 29 orders & 20 orders \\
 & Activities & 55 operations & 44 operations & 18 operations & 139 transports & 348 transports$^{\rm b}$ & 360 transports$^{\rm b}$ \\
 & Resources & 5 machines & 6 work centres & 8 machines, 3 operators & 1 AGV, 1 station & 1 AGV, 1 station & 2 AGVs, 2 robots, buffer of 5 \\
 & Other & 4 product families & -- & 38 eligible pairs & 5 pallets, 5 kit types & 12 pallets, 5 kit types & 18 pallets, 5 kit types \\
\addlinespace[3pt]
\textbf{Variables} & Interval & 55 & 44 & 18 & 168 & 29 & 20 \\
 & Optional interval & -- & -- & 38 & -- & 348 & 1\,480 \\
 & Sequence & 1 & -- & -- & -- & -- & -- \\
 & Integer & -- & 10 & -- & -- & 1\,338 & 1\,502 \\
 & Binary & -- & -- & -- & -- & 348 & 380 \\
 & Total & 56 & 54 & 56 & 168 & 2\,063 & 3\,382 \\
\addlinespace[3pt]
\textbf{Constraints}$^{\rm c}$ & Families & 5 & 4 & 4 & 3 & 9 & 17 \\
 & Total & 60 & 60 & 39 & 141 & 2\,281 & 3\,920 \\
 & Structure & R, P, NW, NO, SD & R, P, NO, L & P, AL, NO, CU & P, NO & P, NO, PR, L & P, NO, NW, AL, CU, PR, L \\
\addlinespace[3pt]
\textbf{Objective} & & $\min C_{\max}$ & $\min \sum_j w_j T_j$ & $\min C_{\max}$ & $\min C_{\max}$ & $\min C_{\max}$ & $\min C_{\max}$ \\
\addlinespace[3pt]
\textbf{Reference} & Value & 1155 min & 5120 & 315 min & 41\,955 s & 13\,815 s & 7\,435 s \\
 & Status & Optimal & Optimal & Optimal & Optimal & Feasible & Feasible \\
 & Gap (\%) & 0.00 & 0.00 & 0.00 & 0.01 & 2.17 & 84.13 \\
\bottomrule
\end{tabularx}
\begin{tabnote}\raggedright
$^{\rm a}$\,FSP: flow shop; JSP: job shop; FJSP: flexible job shop; RCS: resource-constrained scheduling of AGV transports and order picking.
$^{\rm b}$\,Candidate (pallet, order) transports, modelled as optional intervals that exist only if the pallet is assigned to the order.
$^{\rm c}$\,Counted on the reference models, excluding the objective. R: release dates; P: precedence; NW: no-wait or synchronisation; NO: no-overlap; SD: sequence-dependent setups; AL: alternative resources; CU: cumulative resource; PR: presence (assignment) links; L: linear (demand, capacity, tardiness).
\end{tabnote}
\end{table}

Three replicates are run for each combination because LLM output is stochastic: even at the low temperature used here, two generations from the same prompt can differ in their formulation, their tool calls and the resulting script. Each replicate is therefore a fresh generation rather than a repeated execution of the same script.

All runs are evaluated following the same procedure and on the first attempt. The generated script is executed only once, exactly as produced, against the SQLite database associated with the corresponding problem and subject to the 60~s time limit used for the reference models; no run is re-prompted, repaired, or re-solved. Both choices are deliberate.

The 60~s budget is the same as that used to establish the reference results: CP Optimizer proves the optimum for PC1--PC4 within this limit, with solution times ranging from 0.1 to 20~s, and the best-known values for PC5 and PC6 did not improve when their reference models were re-run with a longer time budget. Therefore, a 60-second limit facilitates the execution and comparison of the generated models. In this way, the comparison remains focused on the model rather than on the solver.

The single-attempt evaluation, in turn, makes it possible to isolate the capability under study. Iterative repair loops guided by execution feedback can be an effective complement \citep{ref20_Szeider_CPAgent}; Evaluating only the first attempt makes it possible to assess what each configuration can produce independently, under a fixed and comparable cost, while iterative refinement and reformulation capabilities remain outside the scope of this study.

The metrics derived from this comparison, together with token consumption, number of tool calls, and generation time recorded for each run, are defined in Section~\ref{subsec:metrics}.

\subsection{Evaluation metrics}
\label{subsec:metrics}

Each run is evaluated along the three dimensions of Table~\ref{tab:metrics}. Effectiveness is measured at two levels against the reference objective $z_{\mathrm{ref}}$. The formulation + coding accuracy, $\mathrm{Acc}_{\mathrm{F+C}}$, evaluates the script exactly as generated: it is executed once, without modification, with the database of its problem and the 60~s time limit of the reference models, and counts as correct only if it completes without errors and reports $z = z_{\mathrm{ref}}$. A script that prints $z_{\mathrm{ref}}$ and then fails, for instance while printing the schedule, is therefore incorrect. The formulation accuracy, $\mathrm{Acc}_{\mathrm{F}}$, isolates the model from its implementation. Every run that misses $z_{\mathrm{ref}}$ and has assessable code is reviewed against the reference model, and only its implementation errors are corrected: misuse of the \texttt{docplex.cp} API, including the semantics of optional interval variables, Python and syntax errors, data access and output format. Sets, parameters, variables, constraints and objective are never modified, and the corrected script is re-run under the same conditions. Both accuracies are the mean of the per-run outcome (100\% if correct, 0 otherwise) over the three replicates of a setup, or over all the runs of a group for aggregated values.

\begin{table}[!htb]
\centering
\caption{Evaluation metrics reported for each run.}
\label{tab:metrics}
\footnotesize
\renewcommand{\arraystretch}{1.15}
\begin{tabularx}{\linewidth}{@{}l l Y@{}}
\toprule
Dimension & Metric & Definition \\
\midrule
\begin{tabular}[t]{@{}l@{}}\textbf{Effectiveness}\end{tabular}
  & $\mathrm{Acc}_{\mathrm{F+C}}$ (\%) & Percentage of replications of the same experimental setting for which the generated script executes successfully without modification and returns an objective value satisfying $z = z_{\mathrm{ref}}$ \\
  & $\mathrm{Acc}_{\mathrm{F}}$ (\%)   & Percentage of replications of the same experimental setting that reach $z_{\mathrm{ref}}$ after implementation errors are corrected, without modifying the underlying mathematical formulation \\
\addlinespace[4pt]
\begin{tabular}[t]{@{}l@{}}\textbf{Efficiency}\end{tabular}
  & Tokens     & Input and output tokens of all LLM calls of the run \\
  & Time (s)   & Wall-clock time of the complete workflow \\
  & Tool calls & Number of calls to the MCP documentation server (workflows (a) and (b)) \\
\addlinespace[4pt]
\begin{tabular}[t]{@{}l@{}}\textbf{Diagnostics}\end{tabular}
  & Model outcome        & Result of the generated model, as generated and after correction: reaches $z_{\mathrm{ref}}$; feasible but worse than $z_{\mathrm{ref}}$; invalid (objective below a proven bound); no solution within 60~s or proven infeasible; no objective reported; execution error; no complete script \\
  & First failure point  & Stage of the script at which an incorrect run first fails: output format or syntax, import, model building, data loading or Python runtime, result reporting, or solve without a valid solution \\
  & Implementation errors & Number of distinct coding errors in the script, each classified by nature as a hallucinated symbol (a name that does not exist in \texttt{docplex.cp}), the misuse of an existing API element, a silent semantic error (the script runs but encodes different semantics) or a format or environment defect \\
\bottomrule
\end{tabularx}
\begin{tabnote}\raggedright
The per-run diagnoses, with the individual errors of each script, are given in the complementary tables.
\end{tabnote}
\end{table}

Efficiency is measured by (i) the input and output tokens of all LLM calls of a run, (ii) by the inference time, that is, the wall-clock time of the complete workflow from the reception of the prompt to the final answer and, (iii) by the number of calls to the MCP documentation server. All three are taken from the execution records of Langflow and summarised by their mean and sample standard deviation; five runs of workflow (a) with incomplete token records are excluded from the token statistics.

Finally, every run that is incorrect as generated is diagnosed by the class of its formulation error and by the number and type of its distinct coding errors (Table~\ref{tab:metrics}).

\subsection{Results and Discussion}
\label{subsec:results}

All experiments were run on a laptop (MSI Alpha 17 C7VG) with an AMD Ryzen 9 7945HX processor (16 cores, 32 threads) and 32~GB of DDR5 memory, running Windows 11 (64-bit). The LLMs were queried remotely through the OpenRouter API, so inference times include network latency and provider-side processing. The workflows were orchestrated in Langflow Desktop 1.11.6. The MCP documentation server (MCP Python SDK 1.27.1), the generated scripts and the reference models ran locally with Python~3.14.4, docplex~2.32.264 and IBM ILOG CP Optimizer~20.1.0.

\subsubsection{Overall results}
\label{subsubsec:overall}

Table~\ref{tab:summary} summarises the results of the 54 experimental setups (3 replications per experimental setup). Its most general finding concerns the formulation. Once implementation errors are corrected, the three LLMs formulate the three classical shop problems and the basic AGV problem correctly in practically every run, whatever the workflow: on PC1--PC4, the 91,7\% of the setups are correct in all three replicates. Features that are absent from textbook instances, such as sequence-dependent changeovers, no-wait couplings, machine eligibility and a cumulative pool of operators, are thus translated into correct CP formulations with high reliability. The formulation becomes the limiting factor only in the two extended warehouse problems, PC5 and PC6, whose reference models couple optional transport intervals with integer allocation decisions and more than 2\,000 constraints: there, the formulation accuracy falls below one half in PC5 and to a few isolated runs in PC6.

\begin{table}[p]
\centering
\caption{Summary of results per experimental setup (workflow, problem and LLM): accuracy at both levels, token consumption, inference time and MCP tool calls.}
\label{tab:summary}
{\fontsize{7}{8.4}\selectfont
\setlength{\tabcolsep}{2.6pt}
\renewcommand{\arraystretch}{1.06}
\begin{tabular*}{\linewidth}{@{\extracolsep{\fill}}|c|l|c|r|r|r|r|r|r|}
\hline
\multicolumn{1}{|c|}{} & \multicolumn{1}{c|}{} & \multicolumn{1}{c|}{} & \multicolumn{2}{c|}{Accuracy (\%)} & \multicolumn{2}{c|}{Tokens (k)} & \multicolumn{1}{c|}{Time} & \multicolumn{1}{c|}{Tool} \\
\cline{4-7}
\multicolumn{1}{|c|}{Workflow} & \multicolumn{1}{c|}{Problem} & \multicolumn{1}{c|}{LLM} & F & F+C & Input & Output & (s) & calls \\
\hline
WA & PC1 (FSP) & Gem & \textbf{100.0} & \textbf{100.0} & 152.1 $\pm$ 3.1 & 21.8 $\pm$ 1.5 & 202 $\pm$ 16 & 6.0 $\pm$ 0.0 \\
 &  & GPT & \textbf{100.0} & 66.7 & 414.2 $\pm$ 173.7 & 24.4 $\pm$ 6.2 & 333 $\pm$ 84 & 10.3 $\pm$ 3.1 \\
 &  & DS & \textbf{100.0} & 66.7 & 1108.6 $\pm$ 179.7 & 19.6 $\pm$ 3.2 & 516 $\pm$ 135 & 21.7 $\pm$ 2.5 \\
\cline{2-9}
 & PC2 (JSP) & Gem & \textbf{100.0} & \textbf{100.0} & 140.3 $\pm$ 25.0 & 20.1 $\pm$ 5.0 & 178 $\pm$ 42 & 6.3 $\pm$ 1.2 \\
 &  & GPT & \textbf{100.0} & \textbf{100.0} & 606.7 $\pm$ 121.8 & 26.6 $\pm$ 4.0 & 372 $\pm$ 72 & 13.7 $\pm$ 2.3 \\
 &  & DS & \textbf{100.0} & \textbf{100.0} & 1533.2 $\pm$ 706.3 & 17.9 $\pm$ 2.5 & 442 $\pm$ 42 & 24.3 $\pm$ 5.5 \\
\cline{2-9}
 & PC3 (FJSP) & Gem & \textbf{100.0} & \textbf{66.7} & 149.3 $\pm$ 2.6 & 19.4 $\pm$ 3.7 & 206 $\pm$ 117 & 4.3 $\pm$ 3.8 \\
 &  & GPT & \textbf{100.0} & \textbf{66.7} & 536.6 $\pm$ 147.4 & 31.9 $\pm$ 4.4 & 382 $\pm$ 109 & 13.0 $\pm$ 2.0 \\
 &  & DS & \textbf{100.0} & \textbf{66.7} & 1239.6 $\pm$ 646.4 & 17.2 $\pm$ 3.1 & 404 $\pm$ 78 & 21.0 $\pm$ 7.2 \\
\cline{2-9}
 & PC4 (RCS) & Gem & \textbf{100.0} & \textbf{100.0} & 218.5 $\pm$ 54.4 & 22.9 $\pm$ 5.7 & 242 $\pm$ 42 & 9.0 $\pm$ 1.7 \\
 &  & GPT & \textbf{100.0} & 66.7 & 471.8 $\pm$ 15.5 & 22.7 $\pm$ 1.7 & 319 $\pm$ 159 & 12.0 $\pm$ 1.0 \\
 &  & DS & \textbf{100.0} & 66.7 & 1416.7 $\pm$ 149.6 & 17.4 $\pm$ 1.3 & 435 $\pm$ 73 & 26.3 $\pm$ 2.3 \\
\cline{2-9}
 & PC5 (RCS) & Gem & 33.3 & 33.3 & 147.8 $\pm$ 1.6 & 22.5 $\pm$ 0.4 & 257 $\pm$ 97 & 5.3 $\pm$ 1.2 \\
 &  & GPT & \textbf{100.0} & \textbf{66.7} & 432.6 $\pm$ 52.4 & 28.8 $\pm$ 1.1 & 349 $\pm$ 120 & 11.0 $\pm$ 1.0 \\
 &  & DS & 33.3 & 0.0 & 1158.3 $\pm$ 272.2 & 30.7 $\pm$ 8.2 & 442 $\pm$ 96 & 21.3 $\pm$ 3.2 \\
\cline{2-9}
 & PC6 (RCS) & Gem & \textbf{33.3} & 0.0 & 113.8 $\pm$ 1.0 & 31.0 $\pm$ 12.3 & 320 $\pm$ 97 & 3.7 $\pm$ 3.2 \\
 &  & GPT & 0.0 & 0.0 & 587.0 $\pm$ 273.3 & 48.3 $\pm$ 12.5 & 572 $\pm$ 186 & 13.0 $\pm$ 4.6 \\
 &  & DS & 0.0 & 0.0 & 1240.4 $\pm$ 524.5 & 34.1 $\pm$ 4.4 & 679 $\pm$ 90 & 20.3 $\pm$ 5.7 \\
\hline
WB & PC1 (FSP) & Gem & 66.7 & 0.0 & 99.4 $\pm$ 27.3 & 9.2 $\pm$ 2.9 & 149 $\pm$ 75 & 5.7 $\pm$ 0.6 \\
 &  & GPT & \textbf{100.0} & 33.3 & 406.6 $\pm$ 114.0 & 22.6 $\pm$ 0.6 & 322 $\pm$ 34 & 11.3 $\pm$ 1.5 \\
 &  & DS & \textbf{100.0} & \textbf{66.7} & 724.8 $\pm$ 121.4 & 9.7 $\pm$ 1.8 & 233 $\pm$ 26 & 17.0 $\pm$ 1.7 \\
\cline{2-9}
 & PC2 (JSP) & Gem & \textbf{100.0} & \textbf{66.7} & 109.9 $\pm$ 26.6 & 15.6 $\pm$ 2.6 & 179 $\pm$ 73 & 6.3 $\pm$ 1.2 \\
 &  & GPT & \textbf{100.0} & \textbf{66.7} & 313.3 $\pm$ 71.5 & 16.9 $\pm$ 2.6 & 236 $\pm$ 27 & 10.0 $\pm$ 1.7 \\
 &  & DS & \textbf{100.0} & \textbf{66.7} & 760.8 $\pm$ 239.6 & 7.7 $\pm$ 0.7 & 203 $\pm$ 22 & 17.7 $\pm$ 3.8 \\
\cline{2-9}
 & PC3 (FJSP) & Gem & \textbf{100.0} & \textbf{66.7} & 95.1 $\pm$ 20.2 & 11.9 $\pm$ 1.7 & 124 $\pm$ 18 & 5.7 $\pm$ 0.6 \\
 &  & GPT & \textbf{100.0} & 33.3 & 363.6 $\pm$ 137.5 & 19.0 $\pm$ 2.0 & 271 $\pm$ 25 & 11.0 $\pm$ 2.6 \\
 &  & DS & \textbf{100.0} & 33.3 & 900.8 $\pm$ 79.8 & 10.4 $\pm$ 2.0 & 240 $\pm$ 26 & 19.3 $\pm$ 1.5 \\
\cline{2-9}
 & PC4 (RCS) & Gem & \textbf{100.0} & 33.3 & 89.3 $\pm$ 11.2 & 11.7 $\pm$ 2.3 & 192 $\pm$ 48 & 5.0 $\pm$ 1.0 \\
 &  & GPT & \textbf{100.0} & \textbf{66.7} & 335.7 $\pm$ 40.2 & 15.6 $\pm$ 1.2 & 217 $\pm$ 37 & 10.7 $\pm$ 1.2 \\
 &  & DS & \textbf{100.0} & 33.3 & 667.0 $\pm$ 155.9 & 8.9 $\pm$ 2.0 & 225 $\pm$ 20 & 17.3 $\pm$ 2.1 \\
\cline{2-9}
 & PC5 (RCS) & Gem & 33.3 & 0.0 & 136.4 $\pm$ 21.5 & 16.4 $\pm$ 3.2 & 169 $\pm$ 28 & 6.3 $\pm$ 0.6 \\
 &  & GPT & 33.3 & 0.0 & 332.4 $\pm$ 43.7 & 25.5 $\pm$ 5.0 & 327 $\pm$ 105 & 10.3 $\pm$ 0.6 \\
 &  & DS & \textbf{100.0} & \textbf{33.3} & 1019.7 $\pm$ 300.0 & 14.8 $\pm$ 1.5 & 331 $\pm$ 26 & 20.0 $\pm$ 3.0 \\
\cline{2-9}
 & PC6 (RCS) & Gem & \textbf{33.3} & 0.0 & 73.9 $\pm$ 49.7 & 18.0 $\pm$ 4.1 & 228 $\pm$ 75 & 3.3 $\pm$ 2.1 \\
 &  & GPT & \textbf{33.3} & 0.0 & 382.6 $\pm$ 64.5 & 36.6 $\pm$ 7.6 & 498 $\pm$ 116 & 11.0 $\pm$ 2.0 \\
 &  & DS & 0.0 & 0.0 & 745.9 $\pm$ 315.5 & 26.8 $\pm$ 9.2 & 501 $\pm$ 149 & 16.3 $\pm$ 5.0 \\
\hline
WC & PC1 (FSP) & Gem & \textbf{100.0} & \textbf{33.3} & 2.4 $\pm$ 0.0 & 9.9 $\pm$ 1.2 & 82 $\pm$ 13 & --- \\
 &  & GPT & \textbf{100.0} & 0.0 & 2.2 $\pm$ 0.0 & 9.4 $\pm$ 0.2 & 135 $\pm$ 21 & --- \\
 &  & DS & \textbf{100.0} & 0.0 & 2.3 $\pm$ 0.0 & 7.6 $\pm$ 2.6 & 129 $\pm$ 30 & --- \\
\cline{2-9}
 & PC2 (JSP) & Gem & \textbf{100.0} & 0.0 & 1.9 $\pm$ 0.0 & 9.0 $\pm$ 3.2 & 75 $\pm$ 27 & --- \\
 &  & GPT & \textbf{100.0} & 0.0 & 1.8 $\pm$ 0.0 & 7.2 $\pm$ 0.9 & 91 $\pm$ 16 & --- \\
 &  & DS & \textbf{100.0} & 0.0 & 1.8 $\pm$ 0.0 & 6.9 $\pm$ 6.9 & 142 $\pm$ 160 & --- \\
\cline{2-9}
 & PC3 (FJSP) & Gem & \textbf{100.0} & \textbf{33.3} & 2.0 $\pm$ 0.0 & 9.0 $\pm$ 2.3 & 76 $\pm$ 17 & --- \\
 &  & GPT & \textbf{100.0} & 0.0 & 1.9 $\pm$ 0.0 & 7.2 $\pm$ 0.3 & 99 $\pm$ 20 & --- \\
 &  & DS & 66.7 & \textbf{33.3} & 2.0 $\pm$ 0.0 & 7.9 $\pm$ 6.6 & 195 $\pm$ 191 & --- \\
\cline{2-9}
 & PC4 (RCS) & Gem & \textbf{100.0} & 33.3 & 1.9 $\pm$ 0.0 & 9.4 $\pm$ 2.7 & 110 $\pm$ 78 & --- \\
 &  & GPT & \textbf{100.0} & \textbf{66.7} & 1.8 $\pm$ 0.0 & 7.1 $\pm$ 0.7 & 107 $\pm$ 25 & --- \\
 &  & DS & 66.7 & 33.3 & 1.9 $\pm$ 0.0 & 6.6 $\pm$ 7.3 & 167 $\pm$ 219 & --- \\
\cline{2-9}
 & PC5 (RCS) & Gem & \textbf{66.7} & \textbf{33.3} & 2.4 $\pm$ 0.0 & 13.7 $\pm$ 1.3 & 116 $\pm$ 9 & --- \\
 &  & GPT & 0.0 & 0.0 & 2.2 $\pm$ 0.0 & 11.2 $\pm$ 0.2 & 163 $\pm$ 19 & --- \\
 &  & DS & 0.0 & 0.0 & 2.3 $\pm$ 0.0 & 11.0 $\pm$ 3.2 & 241 $\pm$ 224 & --- \\
\cline{2-9}
 & PC6 (RCS) & Gem & 0.0 & 0.0 & 2.9 $\pm$ 0.0 & 15.0 $\pm$ 0.0 & 161 $\pm$ 65 & --- \\
 &  & GPT & 0.0 & 0.0 & 2.7 $\pm$ 0.0 & 13.3 $\pm$ 1.1 & 191 $\pm$ 15 & --- \\
 &  & DS & \textbf{33.3} & 0.0 & 2.8 $\pm$ 0.0 & 11.1 $\pm$ 6.6 & 272 $\pm$ 197 & --- \\
\hline
\end{tabular*}}
\begin{tabnote}\raggedright Accuracy: percentage of the three replicates of each setup that are correct (100\% for a correct replicate, 0 otherwise); F: formulation accuracy; F+C: formulation + coding accuracy (script as generated). \textbf{Bold}: best LLM for each workflow and problem. Tokens (in thousands), inference time of the complete workflow and MCP tool calls: mean $\pm$ sample standard deviation over the three replicates. WA: multi-agent + MCP; WB: single agent + MCP; WC: direct LLM (no tool calls).
\end{tabnote}
\end{table}

The formulation plus coding accuracy stays below the formulation accuracy in almost every workflow and problem, so the first barrier to a correct result is the implementation of the model rather than the model itself: even on PC1--PC4, about half of the correct formulations are lost in the generated code. The two levels also differ in reproducibility. A correct formulation is usually obtained in all three replicates of a setup, whereas a correct script is not: 84.8\% setups with at least one correct script produce it in only one or two replicates, so a single generation is a weak indicator of the capability of a configuration. The only setups in which the three generated scripts are correct belong to the multi-agent workflow, which reaches up to 67.7\% of  formulation + coding accuracy in PC1-PC4 tests, outperforming the rest of architectures mainly in the coding task.

The LLM has a smaller and less systematic effect on accuracy: the best model changes from one workflow and problem to another, and each of the three LLMs is the most accurate in some setups. Its effect on cost is, in contrast, systematic. Token consumption differs by more than an order of magnitude between the direct LLM and the agentic workflows and, within each agentic workflow, depends mainly on the LLM, whose tool-calling behaviour determines how often the context is re-sent, and only marginally on the problem. Section~\ref{subsubsec:workflow-llm} analyses these effects in detail.

\subsubsection{Workflow and LLM analysis}
\label{subsubsec:workflow-llm}

\begin{figure}[!htb]
\centering
\includegraphics[width=0.49\linewidth]{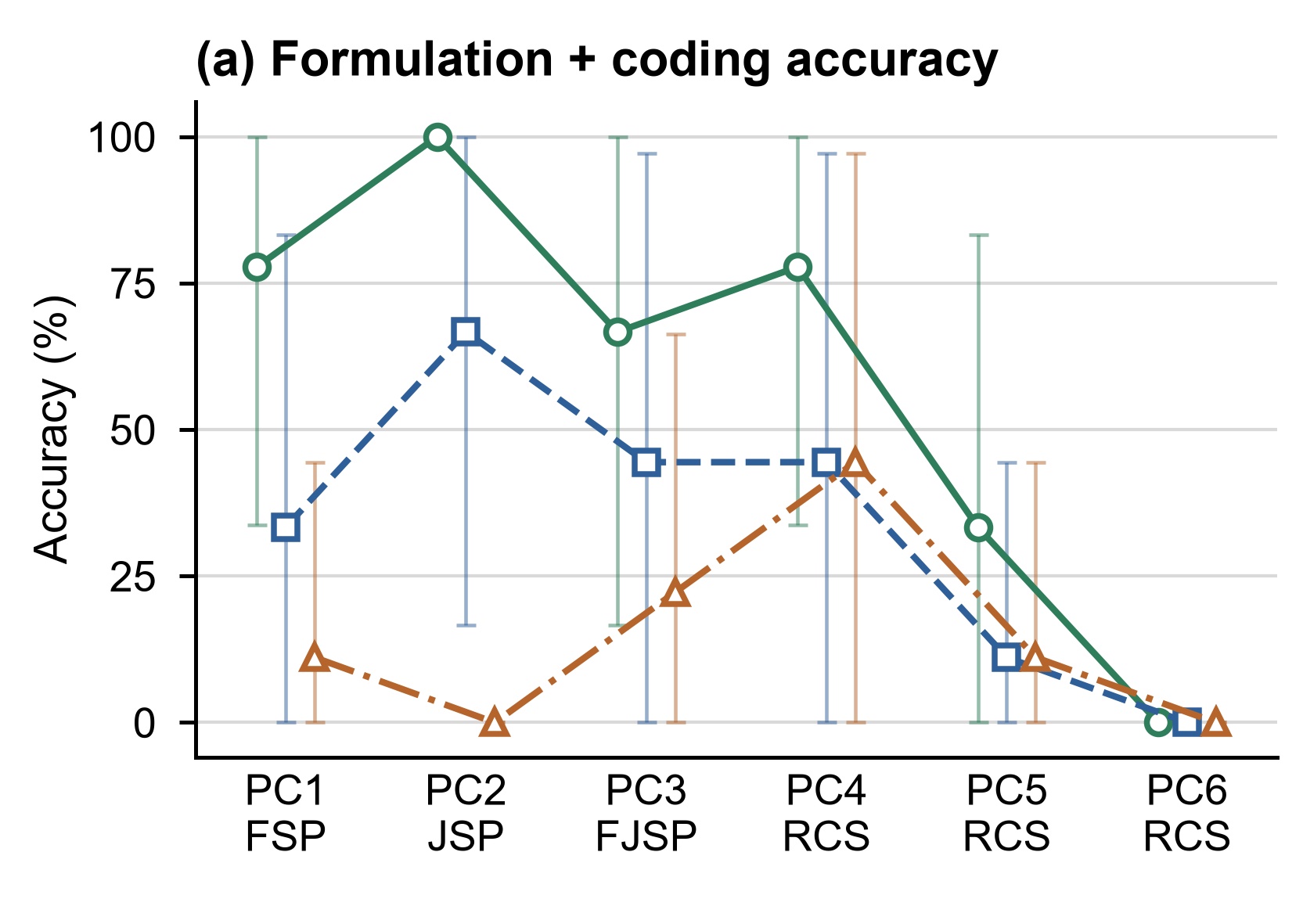}\hfill
\includegraphics[width=0.49\linewidth]{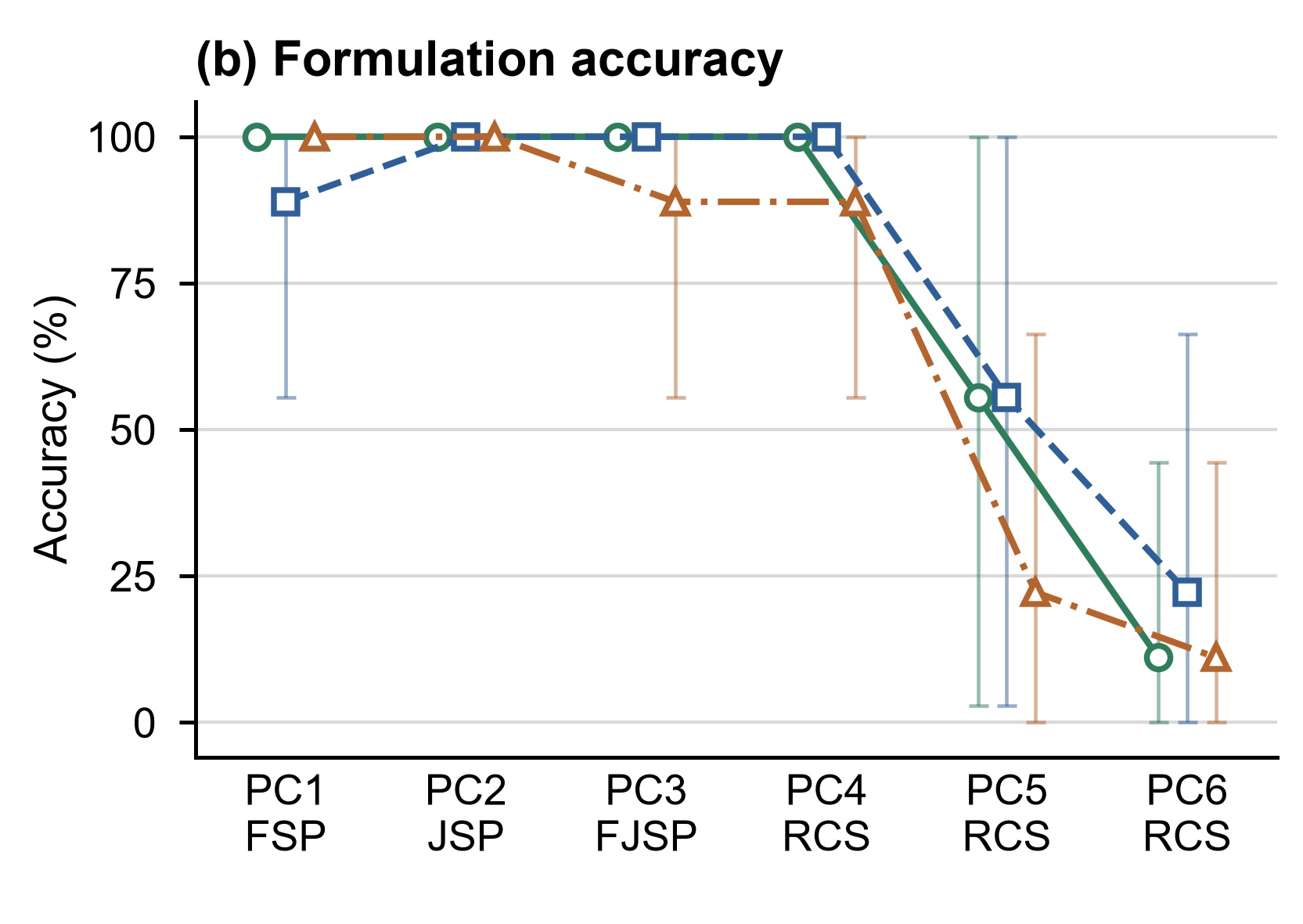}\\[3pt]
\includegraphics[width=0.9\linewidth]{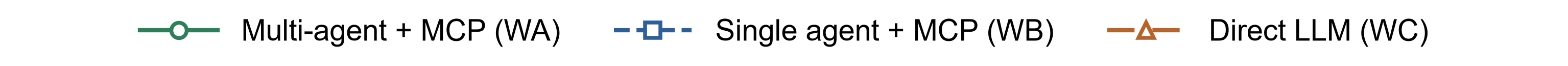}
\caption{Accuracy per problem and workflow across the three LLMs: (a) formulation + coding accuracy, $\mathrm{Acc}_{\mathrm{F+C}}$; (b) formulation accuracy, $\mathrm{Acc}_{\mathrm{F}}$. Mean $\pm$ standard deviation over the nine runs of each workflow and problem (three LLMs $\times$ three replicates).}
\label{fig:accuracy-workflows}
\end{figure}

\begin{figure}[!t]
\centering
\includegraphics[width=0.325\linewidth]{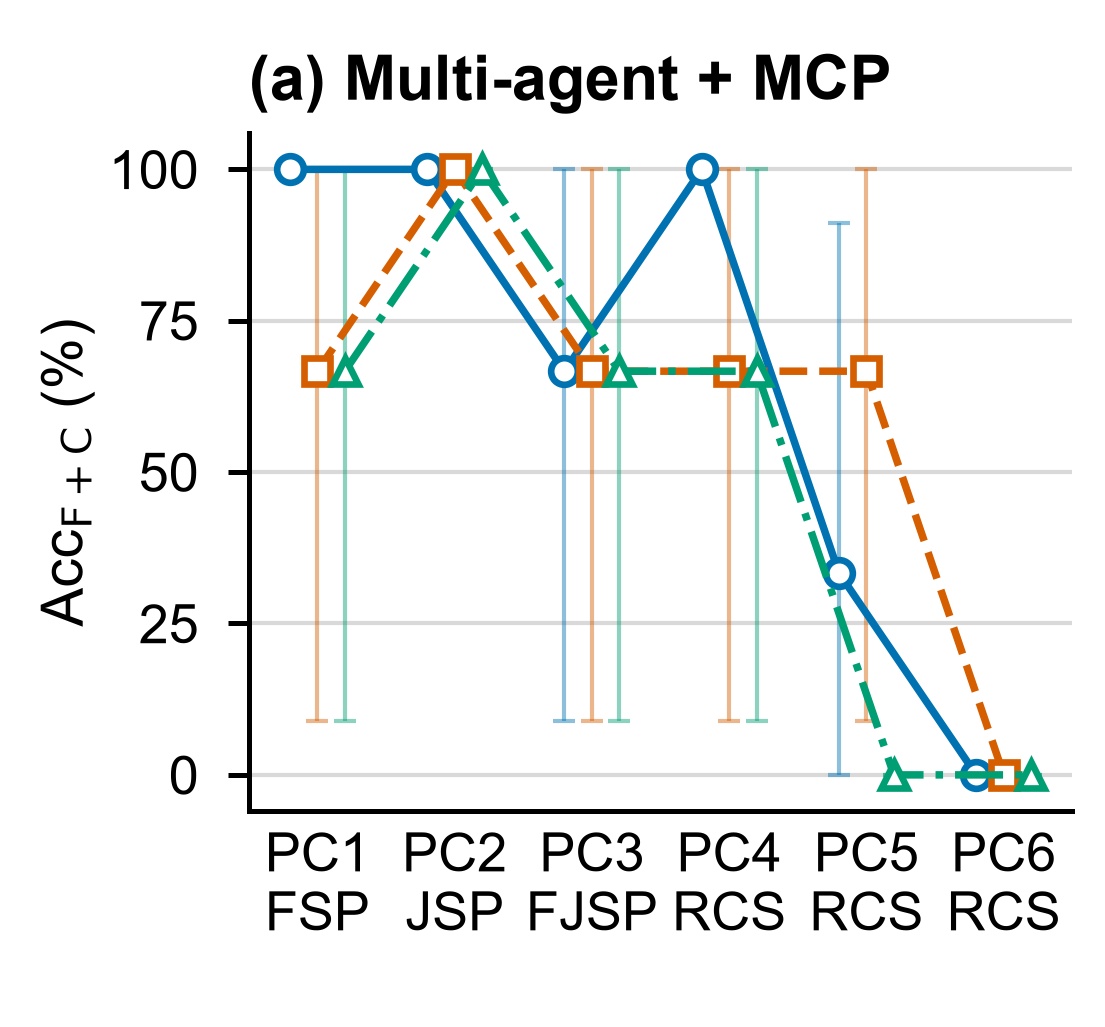}\hfill
\includegraphics[width=0.325\linewidth]{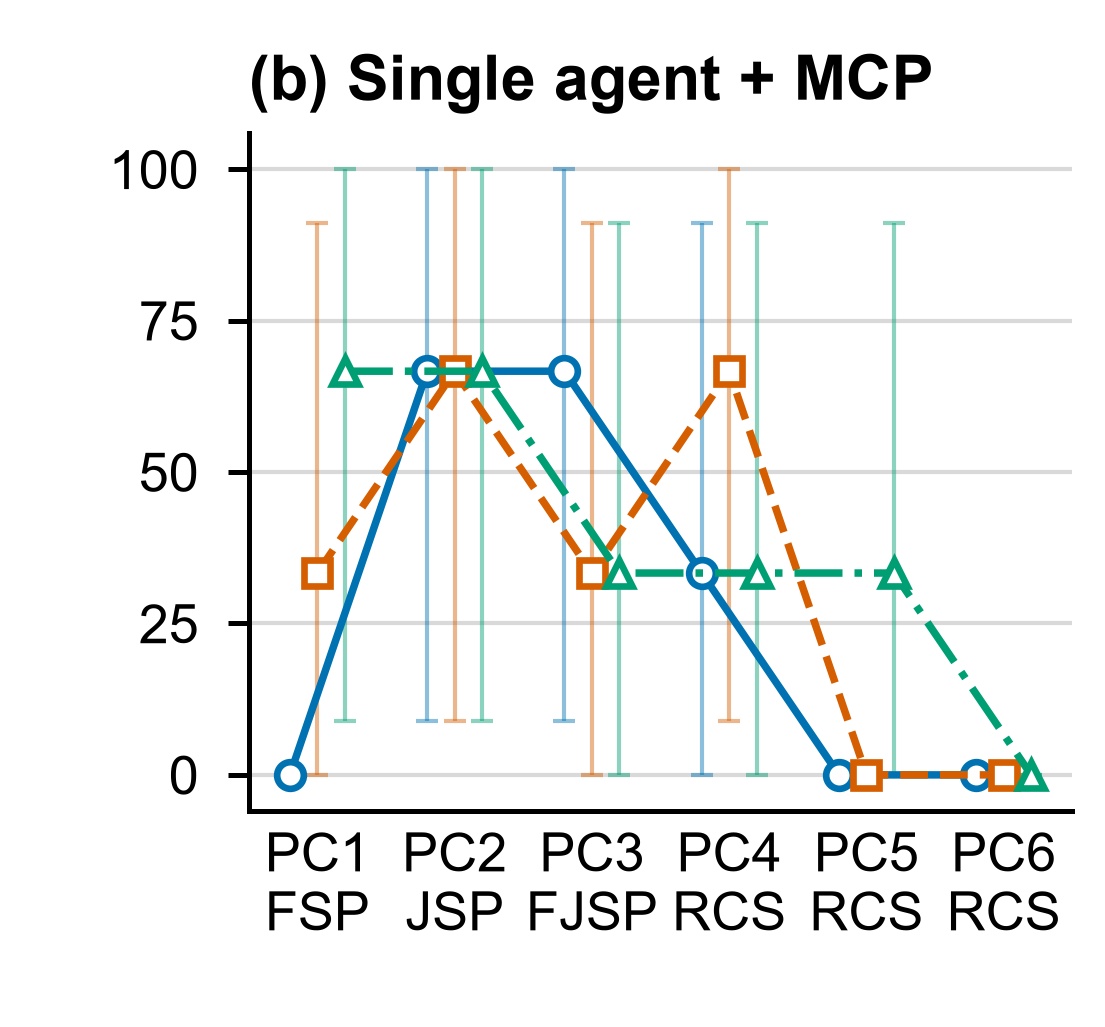}\hfill
\includegraphics[width=0.325\linewidth]{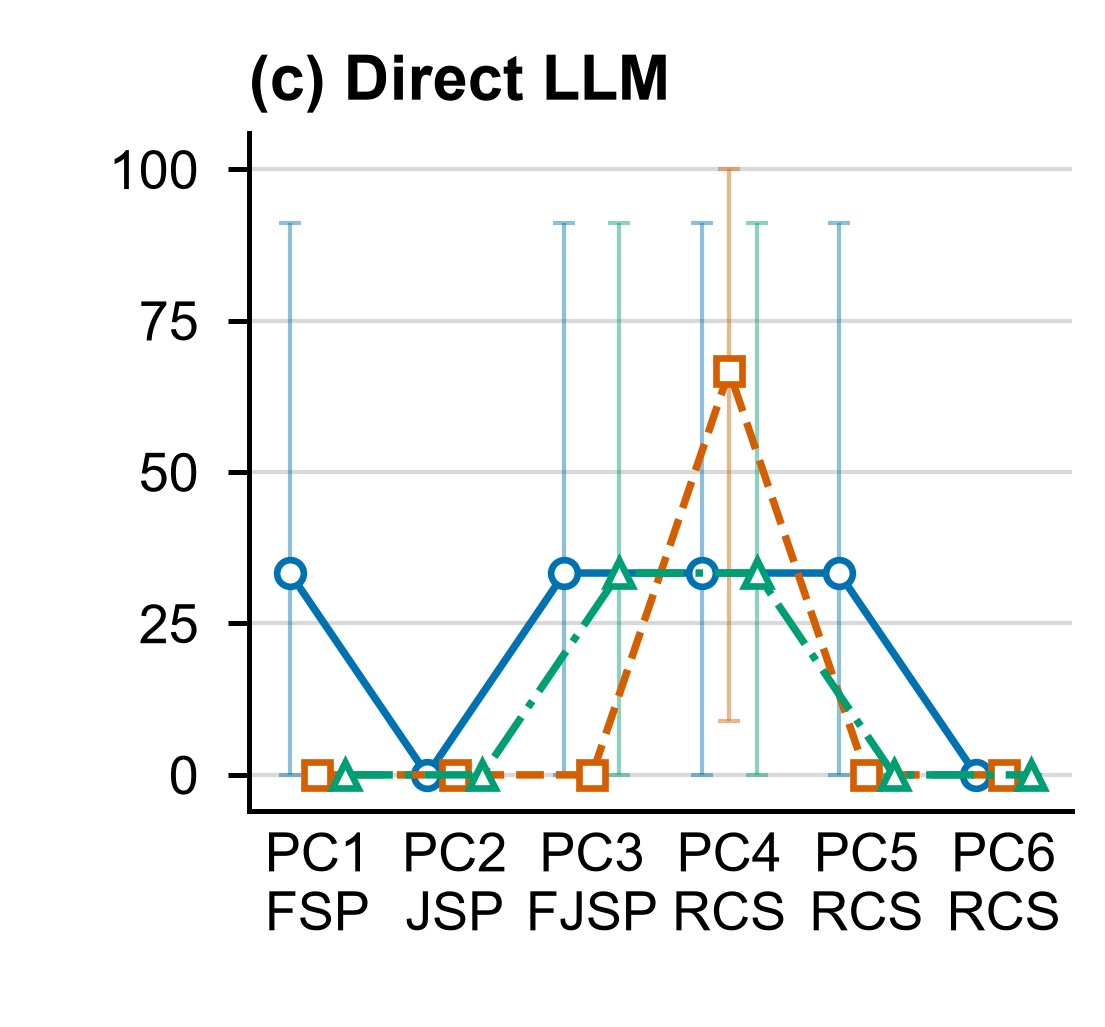}\\[3pt]
\includegraphics[width=0.62\linewidth]{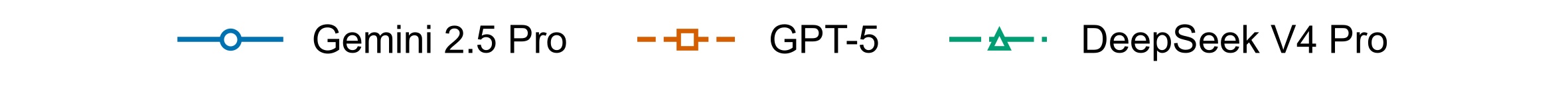}
\caption{Formulation + coding accuracy, $\mathrm{Acc}_{\mathrm{F+C}}$, per problem, LLM and workflow: mean $\pm$ standard deviation over three replicates.}
\label{fig:accuracy}
\end{figure}

Figure~\ref{fig:accuracy-workflows} compares the three workflows at both accuracy levels, aggregating the three LLMs. The agentic configurations improve mainly the implementation, specifically the multi-agent approach. The formulation + coding accuracy rises from 14.8\% with the direct LLM to 33.3\% with the single agent and 59.3\% with the multi-agent workflow, and the share of runs lost only to implementation errors, $\mathrm{Acc}_{\mathrm{F}} - \mathrm{Acc}_{\mathrm{F+C}}$, falls from 53.7 to 44.4 and 18.5 percentage points, respectively. The gain holds for every LLM, each of which reaches its highest formulation + coding accuracy in workflow (a) (Figure~\ref{fig:accuracy}). The formulation accuracy, in contrast, hardly depends on the configuration (Figure~\ref{fig:accuracy-workflows}b): it is 77.8\% in both agentic workflows and 68.5\% with the direct LLM.

\begin{figure}[!t]
\centering
\includegraphics[width=0.325\linewidth]{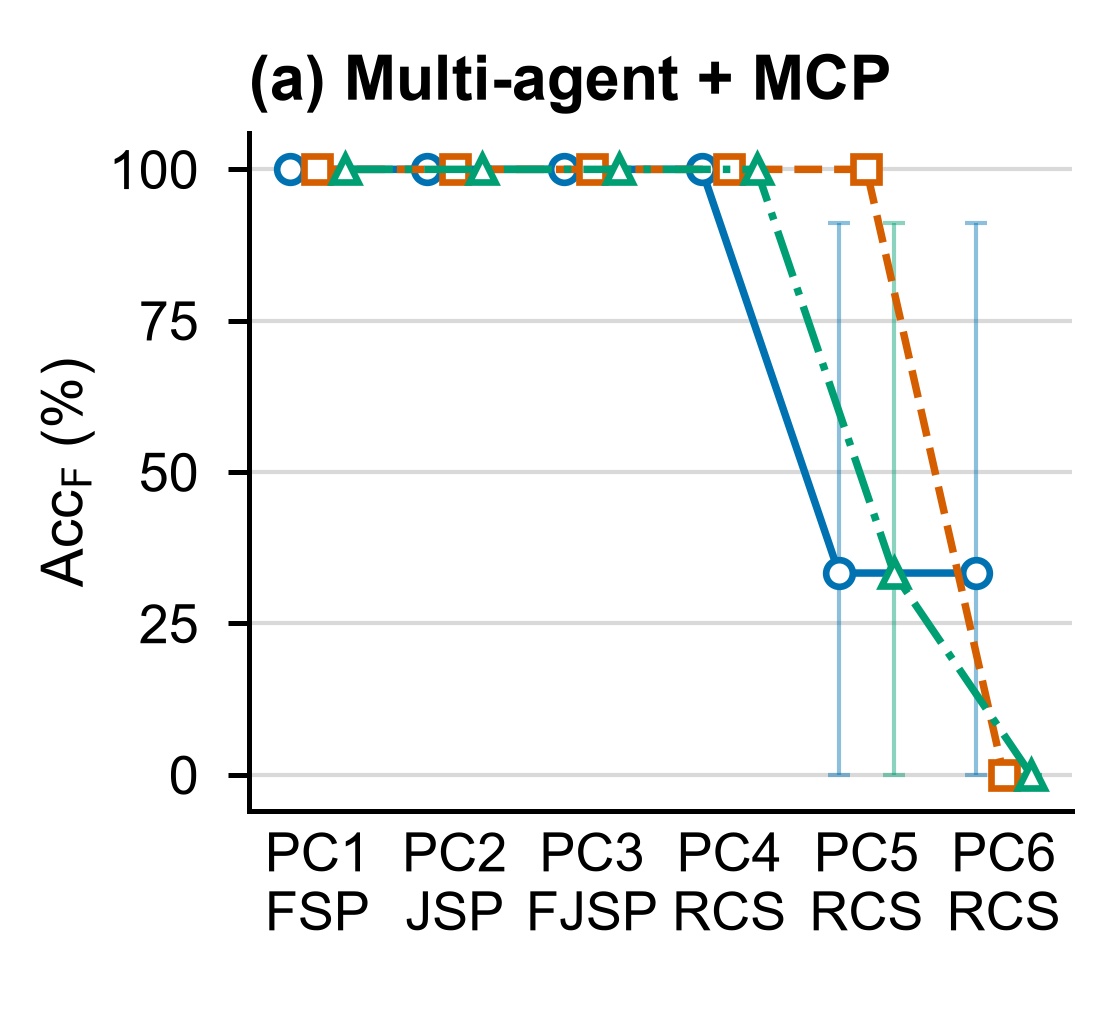}\hfill
\includegraphics[width=0.325\linewidth]{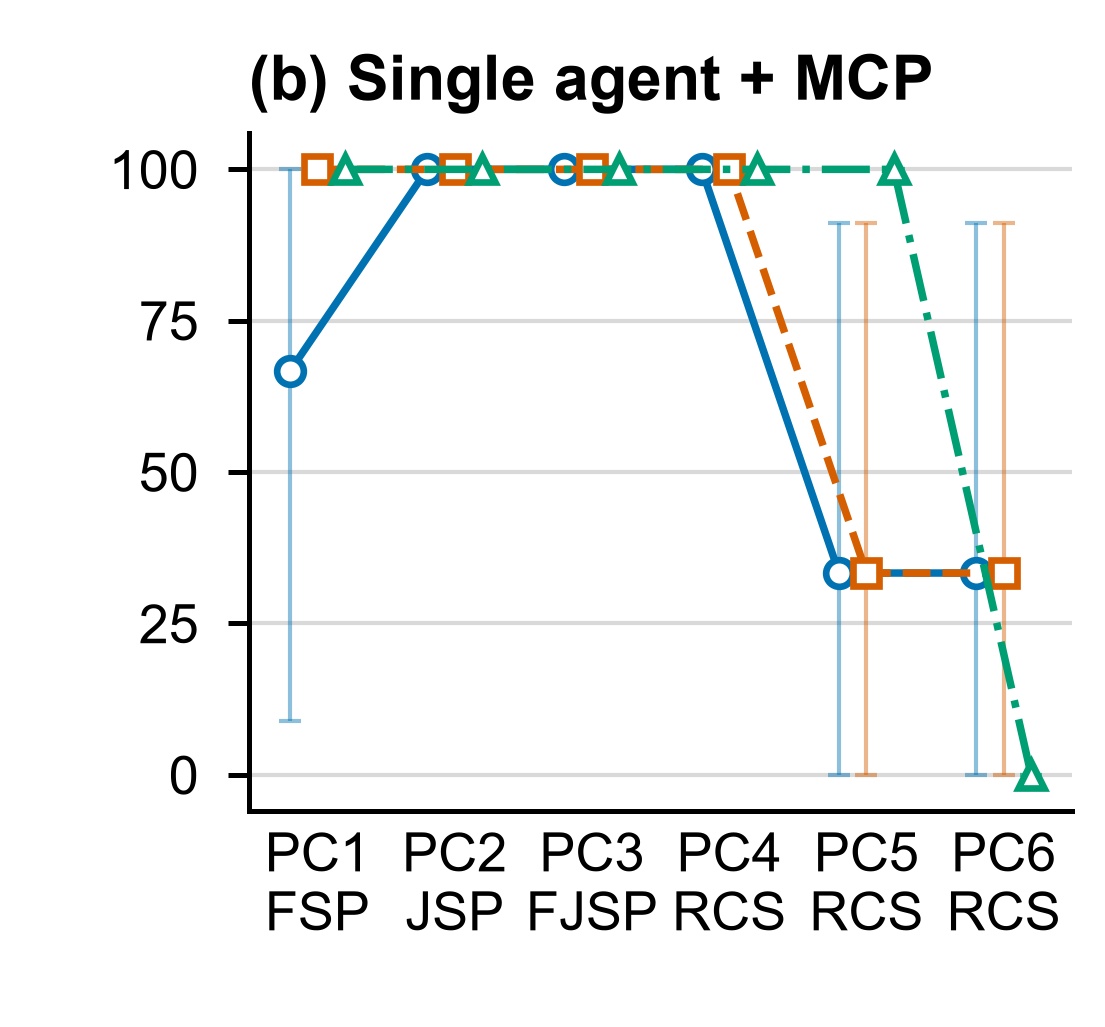}\hfill
\includegraphics[width=0.325\linewidth]{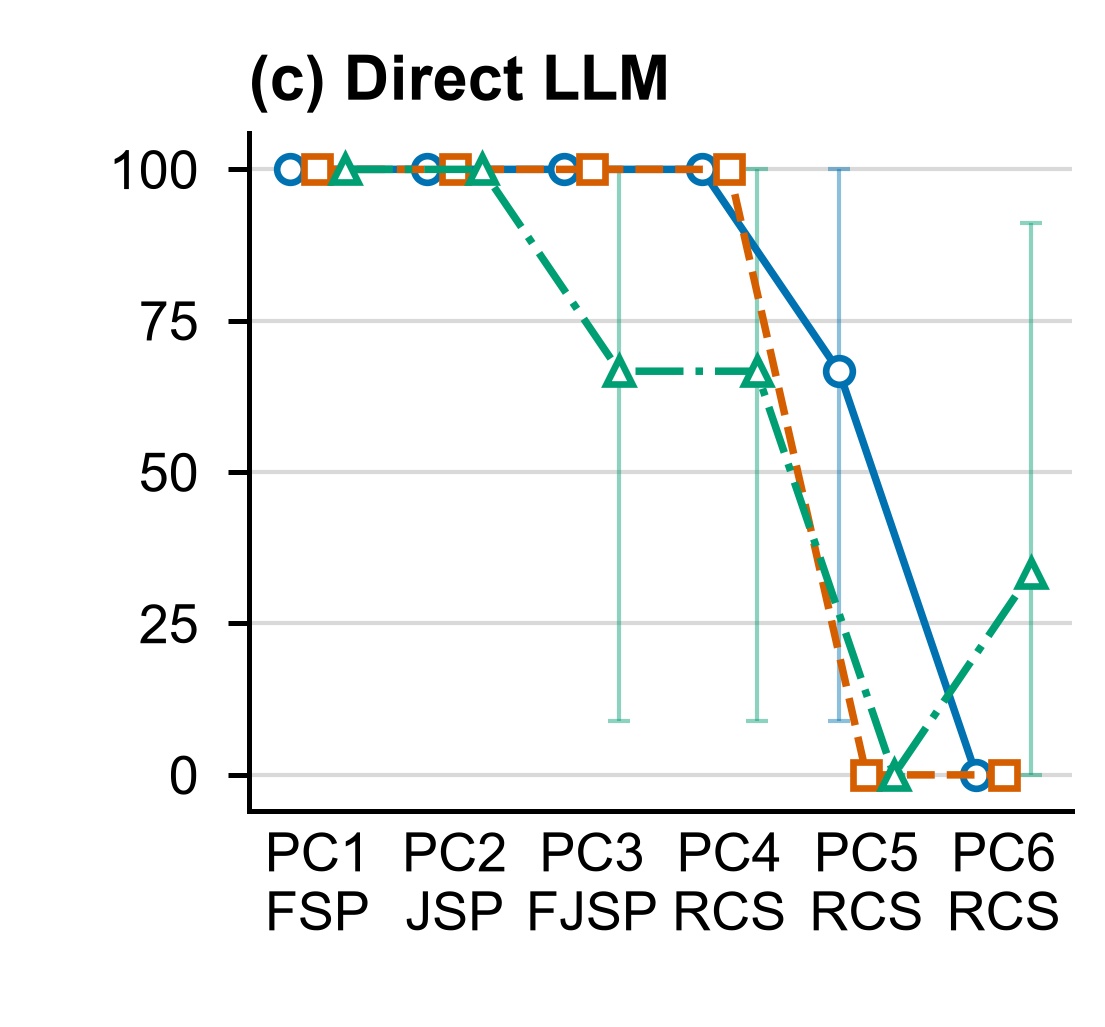}\\[3pt]
\includegraphics[width=0.62\linewidth]{figures/fig_accuracy_legend.jpg}
\caption{Formulation accuracy, $\mathrm{Acc}_{\mathrm{F}}$, per problem, LLM and workflow: mean $\pm$ standard deviation over three replicates.}
\label{fig:accuracy-F}
\end{figure}

Figures~\ref{fig:accuracy} and~\ref{fig:accuracy-F} break both accuracies down by LLM. At the formulation + coding level the ranking of the LLMs changes with the workflow: Gemini~2.5~Pro is the most accurate model in workflows (a) and (c), with 66.7\% and 22.2\%, and DeepSeek~V4~Pro in workflow (b), with 38.9\%. These differences, at most 16.7 points within a workflow, are much smaller than those between workflows. At the formulation level the LLMs are closer still, between 72.2 and 83.3\% in the agentic workflows and between 61.1 and 77.8\% with the direct LLM. On PC1--PC4, every LLM reaches $z_{\mathrm{ref}}$ after implementation correction in every run with assessable code, so the differences come from PC5 and PC6: GPT-5 formulates PC5 correctly in the three replicates of workflow (a), and DeepSeek~V4~Pro in those of workflow (b), whereas PC6 reaches $z_{\mathrm{ref}}$ in only 4 of its 27 runs. The two levels do not rank the LLMs in the same way either. In workflow (a), GPT-5 has the highest formulation accuracy, 83.3\%, but produces fewer correct scripts than Gemini~2.5~Pro, 61.1 against 66.7\%, which indicates that the reliability of the implementation is a property of the LLM distinct from its modelling capability.

\begin{figure}[!htb]
\centering
\includegraphics[width=0.325\linewidth]{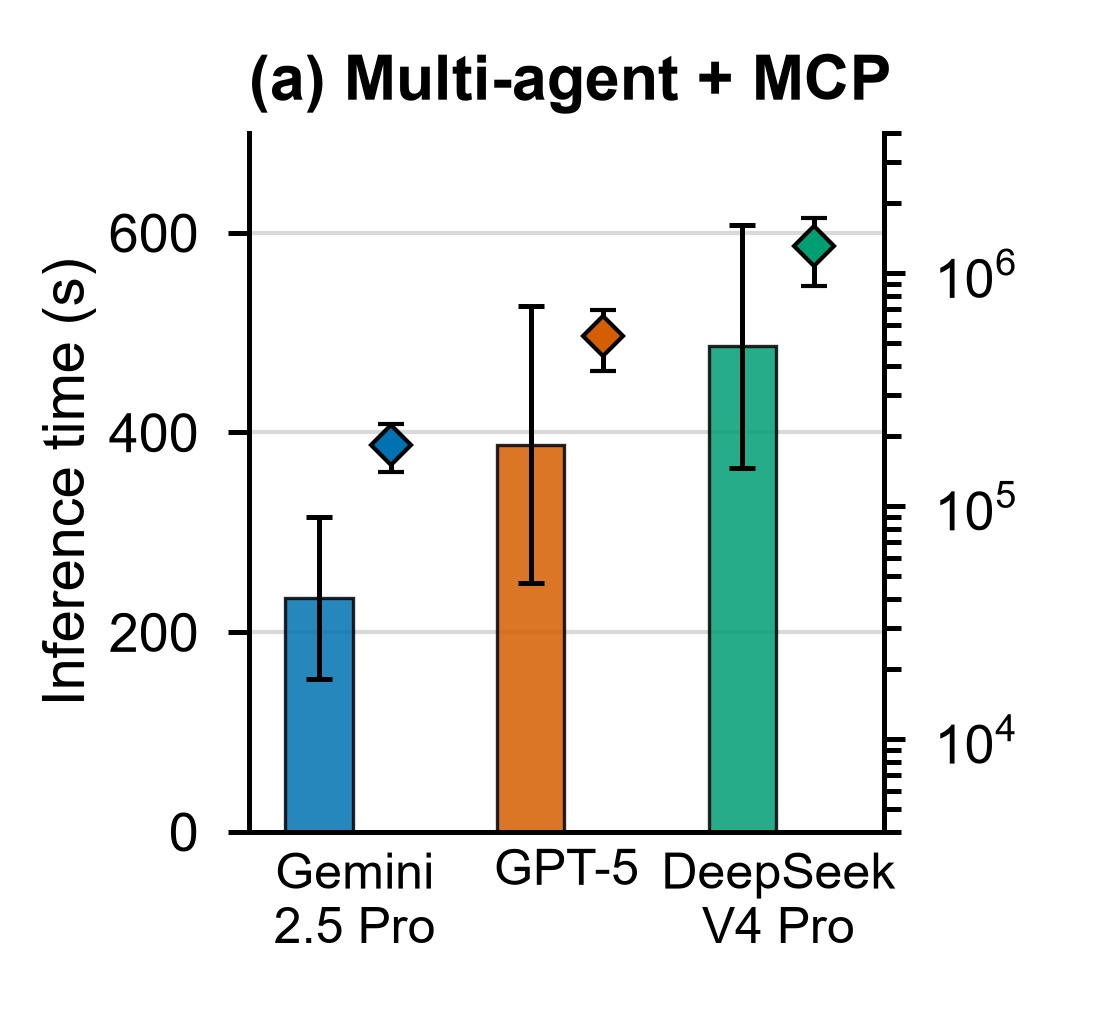}\hfill
\includegraphics[width=0.325\linewidth]{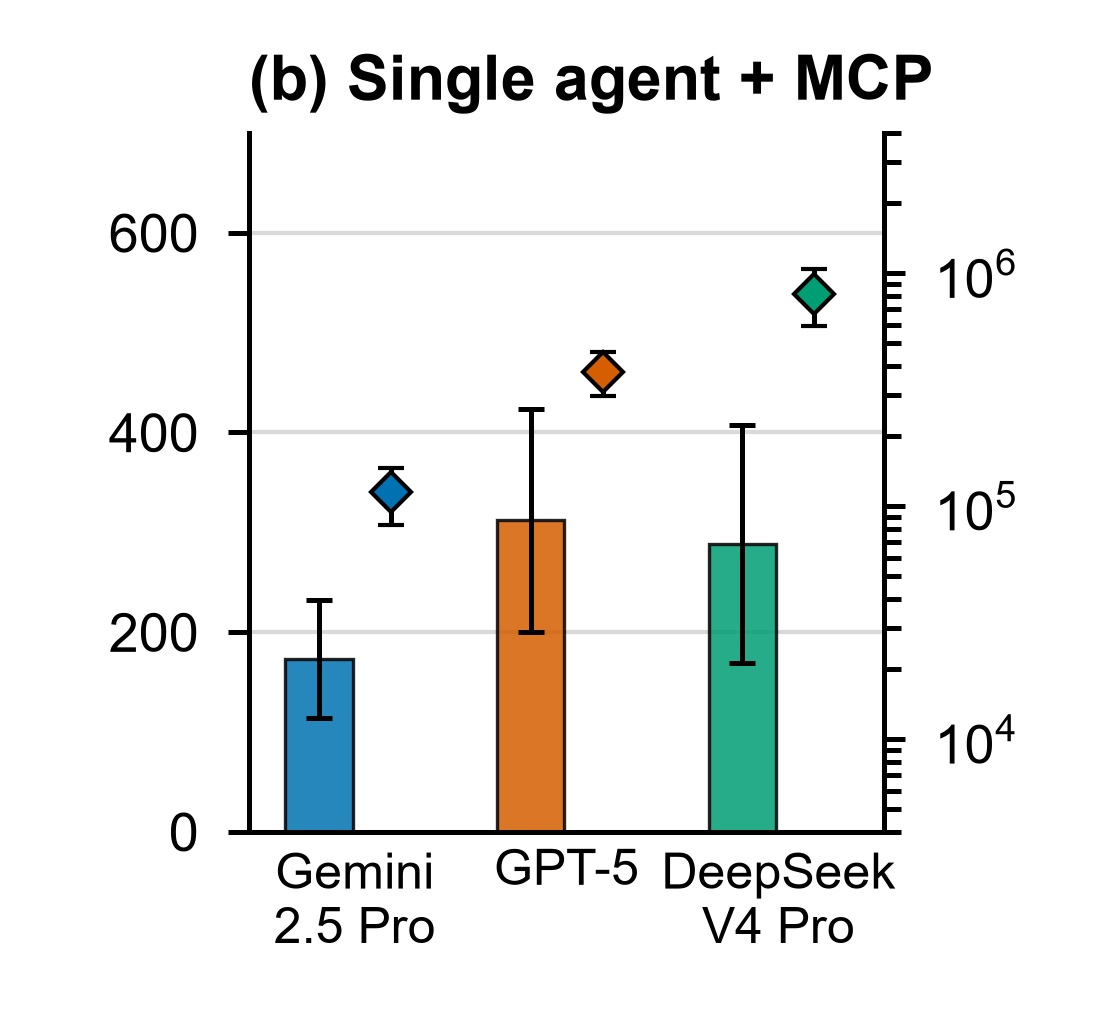}\hfill
\includegraphics[width=0.325\linewidth]{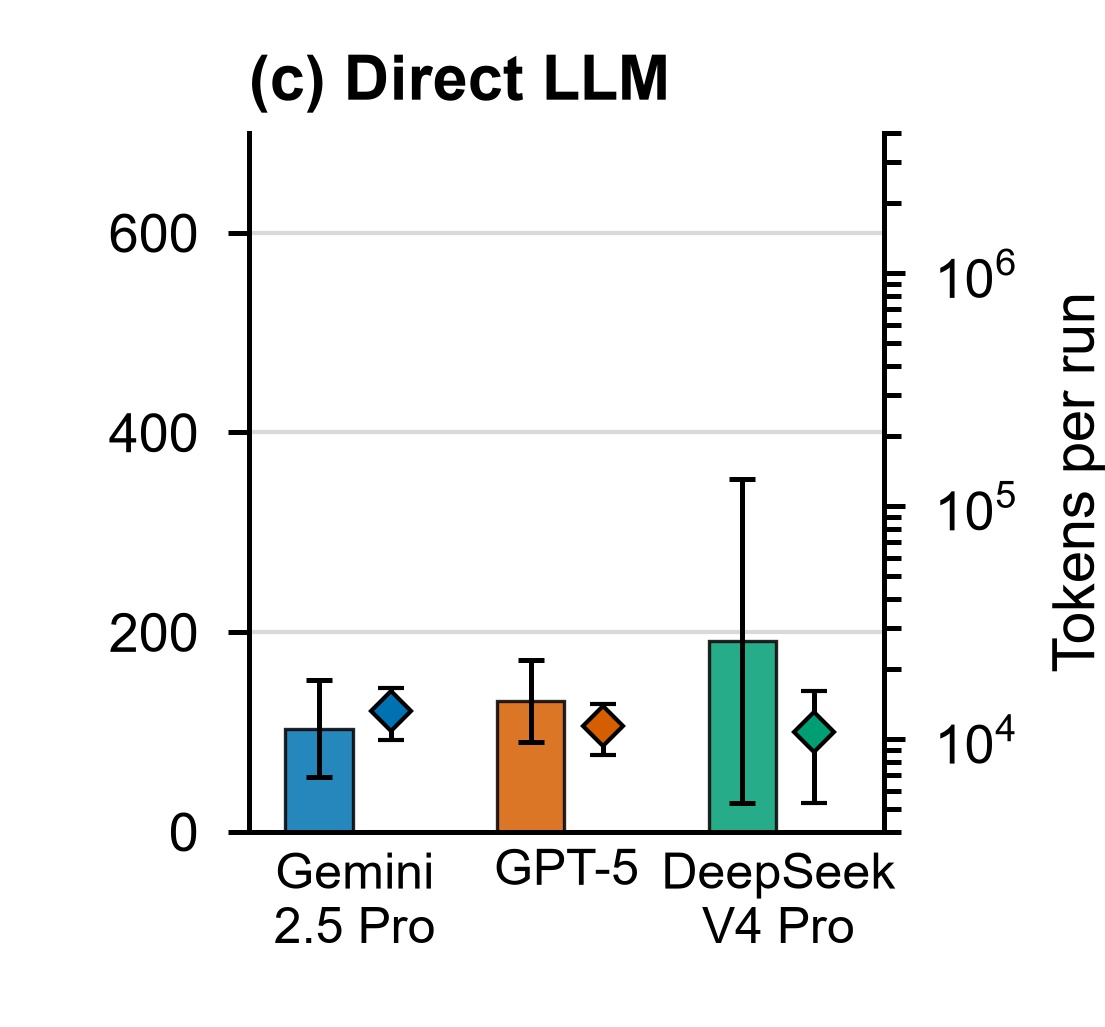}\\[3pt]
\includegraphics[width=0.72\linewidth]{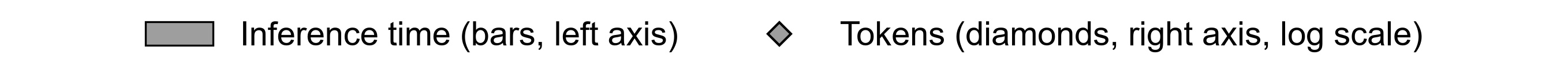}
\caption{Inference time (bars, left axis) and tokens per run (diamonds, right axis, logarithmic scale) per LLM and workflow: mean $\pm$ standard deviation over the 18 runs of each workflow--LLM pair.}
\label{fig:efficiency}
\end{figure}

Figure~\ref{fig:efficiency} and Table~\ref{tab:summary} show the cost of these results. The agentic workflows are considerably more expensive than the direct LLM: workflow (a) requires on average 370~s and 0.73 million tokens per run and workflow (b) 258~s and 0.44 million tokens, against 142~s and about 12\,000 tokens, mostly output tokens, for workflow (c). Workflow (a) thus needs 43\% more time and 66\% more tokens than workflow (b), in exchange for 26 more points of formulation + coding accuracy. In both agentic workflows, 96\% of the tokens are input tokens, and their number grows almost linearly with the number of tool calls, because the whole conversation, including the retrieved documentation, is re-sent at every turn of the tool-calling loop. The Formulator Agent accounts for only 1.7\% of the tokens of workflow (a), but for 30\% of its time, since it writes a long formulation in a single call. Within each workflow, the cost depends on the LLM far more than on the problem. DeepSeek~V4~Pro consumes about seven times more tokens than Gemini~2.5~Pro in both agentic workflows (1.31 against 0.18 million tokens per run in workflow (a) and 0.82 against 0.11 million in workflow (b)), since it makes about four times more tool calls in (a) and three times more in (b), with GPT-5 in between, whereas the problem changes the consumption of a given workflow--LLM pair by less than a factor of 1.8. The inference time also depends on the LLM, but not in proportion to the tokens: GPT-5, which writes the longest outputs, is the slowest model in workflow (b), although it uses less than half the tokens of DeepSeek~V4~Pro. In workflow (a), Gemini~2.5~Pro is both the cheapest and the most accurate model.

\subsubsection{Generated Models and Error Analysis}
\label{subsubsec:generated-model-results}

Table~\ref{tab:outcomes} classifies the outcome of the 162 generated models before and after the correction of their implementation errors. As generated, 58 scripts return the reference objective, 81 fail with an execution error, 14 of them after having printed $z_{\mathrm{ref}}$, six answers contain no complete script, and only 17 run to completion with a wrong or missing result: 14 find no solution within 60~s or prove their model infeasible, one reports an objective below the provable bound of PC5, two report no objective and two answers contain no code. No script delivers a feasible schedule worse than the reference: when a generated model runs, it either reproduces $z_{\mathrm{ref}}$ or fails to solve. After the minimal correction of implementation errors, 121 models reach $z_{\mathrm{ref}}$, seven return feasible but worse schedules, between 2 and 8\% above $z_{\mathrm{ref}}$ in PC5 and twice $z_{\mathrm{ref}}$ in a single PC6 run, and 27 still find no solution within 60~s. The difference between the two blocks of the table is the implementation. On PC1--PC4 it closes almost completely, from 53 to 105 correct models out of 108, whereas on PC5 and PC6 it only moves from 5 to 16 of 54, because the remaining models are over-constrained or structurally wrong. The workflow determines where a run is lost: the direct LLM loses 38 of its 54 scripts to execution errors and the single agent 29, against 14 for the multi-agent workflow, the only one in which solver-level outcomes are as frequent as crashes.
\begin{table}[!htb]
\centering
\caption{Outcome of the 162 generated models, as generated and after correcting their implementation errors, aggregated per workflow and per problem.}
\label{tab:outcomes}
{\footnotesize
\setlength{\tabcolsep}{3.5pt}
\renewcommand{\arraystretch}{1.08}
\begin{tabularx}{\linewidth}{@{}>{\raggedright\arraybackslash}X*{7}{r}@{}}
\toprule
 & \multicolumn{3}{c}{Workflow} & \multicolumn{3}{c}{Problem} & \\
\cmidrule(lr){2-4}\cmidrule(lr){5-7}
Outcome of the generated model & (a) & (b) & (c) & PC1--PC4 & PC5 & PC6 & All \\
Runs & 54 & 54 & 54 & 108 & 27 & 27 & 162 \\
\midrule
\multicolumn{8}{@{}l}{\textit{Script as generated (single execution, no edits)}} \\[1pt]
Reaches $z_{\mathrm{ref}}$ (correct model) & 32 & 18 & 8 & 53 & 5 & -- & 58 \\
Invalid: objective below a proven bound & -- & -- & 1 & -- & 1 & -- & 1 \\
No solution within 60~s or proven infeasible & 7 & 5 & 2 & -- & 4 & 10 & 14 \\
Completes without reporting an objective & 1 & 1 & -- & 1 & 1 & -- & 2 \\
Script fails with an error & 14 & 29 & 38 & 51 & 16 & 14 & 81 \\
\quad of which after printing $z_{\mathrm{ref}}$ & 2 & 3 & 9 & 12 & 2 & -- & 14 \\
No complete script (not assessable) & -- & 1 & 5 & 3 & -- & 3 & 6 \\
\addlinespace[1pt]
$\mathrm{Acc}_{\mathrm{F+C}}$ (\%) & 59.3 & 33.3 & 14.8 & 49.1 & 18.5 & 0.0 & 35.8 \\
\midrule
\multicolumn{8}{@{}l}{\textit{After correcting implementation errors only (formulation unchanged)}} \\[1pt]
Reaches $z_{\mathrm{ref}}$ (correct model) & 42 & 42 & 37 & 105 & 12 & 4 & 121 \\
Feasible schedule, worse than $z_{\mathrm{ref}}$ & 3 & 3 & 1 & -- & 6 & 1 & 7 \\
Invalid: objective below a proven bound & -- & -- & 1 & -- & 1 & -- & 1 \\
No solution within 60~s or proven infeasible & 9 & 8 & 10 & -- & 8 & 19 & 27 \\
No complete script (not assessable) & -- & 1 & 5 & 3 & -- & 3 & 6 \\
\addlinespace[1pt]
$\mathrm{Acc}_{\mathrm{F}}$ (\%) & 77.8 & 77.8 & 68.5 & 97.2 & 44.4 & 14.8 & 74.7 \\
\bottomrule
\end{tabularx}}
\begin{tabnote}\raggedright Counts of runs. As generated, a run reaches $z_{\mathrm{ref}}$ only if the script completes without error and reports $z = z_{\mathrm{ref}}$ (strict criterion); (a) multi-agent + MCP; (b) single agent + MCP; (c) direct LLM. $\mathrm{Acc}_{\mathrm{F+C}}$ and $\mathrm{Acc}_{\mathrm{F}}$: share of runs reaching $z_{\mathrm{ref}}$ at each level (Table~\ref{tab:metrics}).
\end{tabnote}
\end{table}

\vspace{5pt}

\begin{figure}[!htb]
\centering
\includegraphics[width=0.96\linewidth]{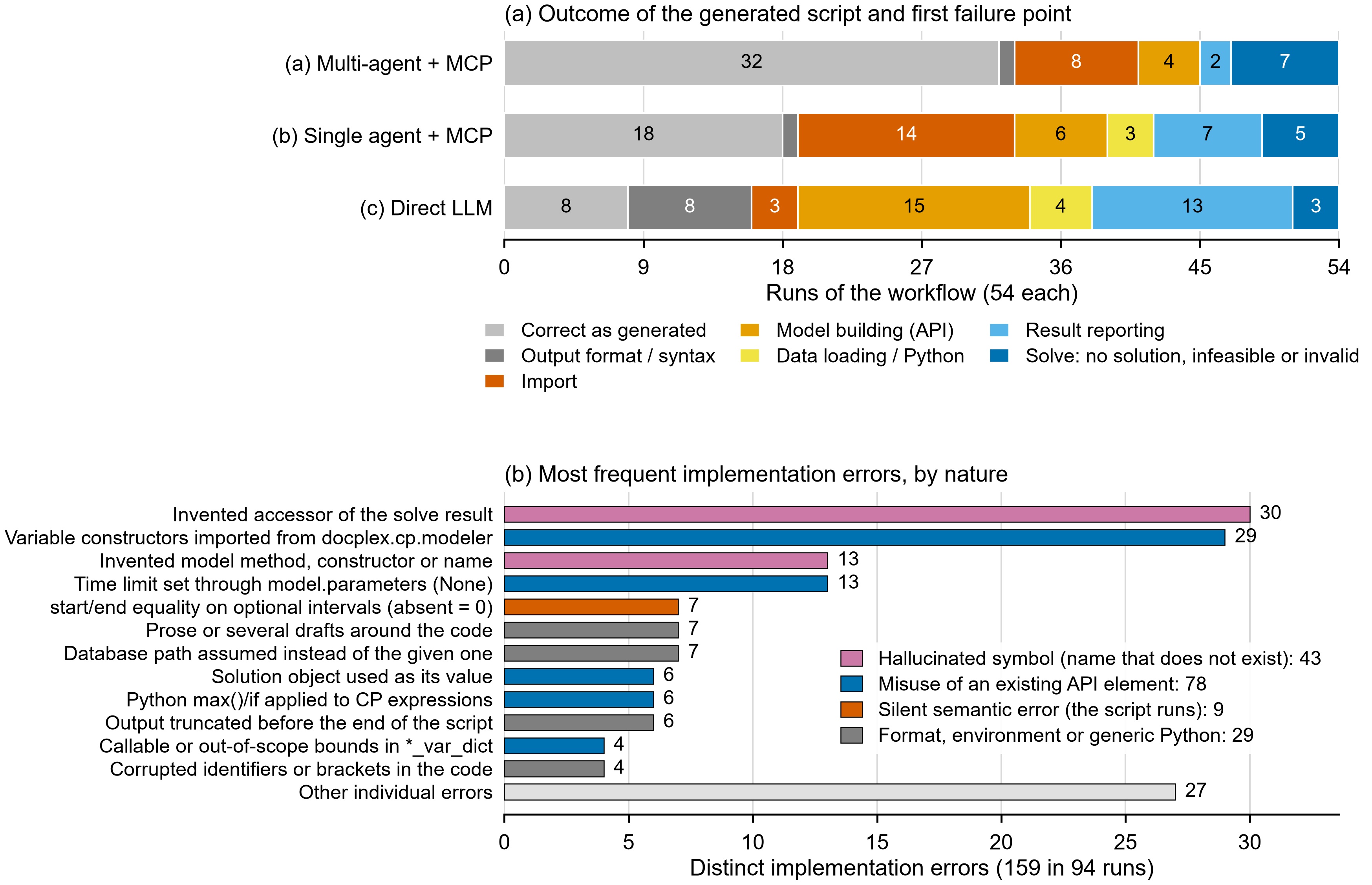}
\caption{Error analysis of the generated scripts. (a) Outcome of the 54 scripts of each workflow: correct as generated, or first point at which the script fails; result reporting includes the 14 scripts that fail after printing $z_{\mathrm{ref}}$, and the last class groups the scripts that run to completion without a valid solution. (b) The twelve most frequent implementation errors among the 159 distinct errors of the 104 incorrect runs, grouped by nature: hallucinated symbol (a name that does not exist in \texttt{docplex.cp}), misuse of an existing element of the API, silent semantic error, and format, environment or generic Python defects.}
\label{fig:error-analysis}
\end{figure}

Figure~\ref{fig:error-analysis} details why the 104 incorrect runs fail (those that did not reach $z_{\mathrm{ref}}$ at first). Panel (a) locates, for each of the 54 scripts of a workflow, the first point at which it fails, ordered as the stages of the pipeline. The three workflows fail at different depths. In the agentic workflows, the most frequent first failure is the import stage, before any model exists (8 runs in workflow (a) and 14 in workflow (b), against 3 with the direct LLM), followed by the model-building calls (4 and 6 runs). With the direct LLM the failures move downstream: 15 scripts fail while building the model and 13 while reporting a model that had already been solved, 9 of them after printing $z_{\mathrm{ref}}$ (Table~\ref{tab:outcomes}), and 8 answers are not even valid Python because prose, several drafts or corrupted text surround the code, a class that is practically absent from the agentic workflows. Only the last class of the panel does not correspond to an interpreter error: 15 scripts (7, 5 and 3) reach the solver and end without a valid solution, and these are the runs in which the formulation, not the code, is at fault. Two consequences follow for the modelling work that a practitioner would face after a generation. First, the failures are overwhelmingly loud: 89 of the 104 incorrect scripts stop with an explicit error message or an unusable output, whereas the silent failures that produce a wrong schedule without any warning are confined to the last class. Second, they are recoverable: once the implementation errors are corrected with the minimal literal patch, 63 of the 104 scripts return $z_{\mathrm{ref}}$ (121 correct models against 58 in Table~\ref{tab:outcomes}), and the remaining ones do not fail any more but expose a formulation problem, as the increase of the no-solution and worse-schedule outcomes in the second block of the table shows.

Panel (b) classifies the 159 distinct implementation errors found in the 94 runs with coding errors by their nature, and separates two mechanisms that are often conflated under the term hallucination. Only 43 errors, roughly one in four, are hallucinations in the strict sense, that is, calls to symbols that do not exist in \texttt{docplex.cp}: 30 of them are invented accessors of the solve result and 13 are invented model methods, constructors or names. The largest group, 78 errors, is instead the misuse of an element that does exist: the variable constructors imported from \texttt{docplex.cp.modeler} instead of \texttt{docplex.cp.model} (29), the time limit assigned to \texttt{model.parameters}, an attribute that is not initialised on a new model (13), a solution object used as if it were its value (6), Python's \texttt{max()} and \texttt{if} applied to CP expressions (6), or callable bounds passed to the variable-dictionary constructors (4). The two mechanisms also leave different traces in panel (a): the import idiom stops the script before the model is built, which is the dominant first failure of the agentic workflows, whereas an invented accessor is only reached after a successful solve, which is why the reporting stage concentrates the failures of the direct LLM. The remaining errors are 29 format and environment defects, of which prose around the code (7), an assumed database path (7), output truncated before the end of the script (6) and corrupted identifiers (4) are the most frequent, and 9 silent errors, 7 of them of the same kind: equating the start or end of an optional interval to another time, which evaluates to zero when the interval is absent and turns a correct PC6 model into an infeasible one without any error message. The 27 further errors grouped as other individual errors are of more than twenty different kinds, none occurring more than three times, so the corpus of implementation errors is highly concentrated: five recurrent mistakes explain more than half of it, which is precisely the situation in which a documentation server can be effective.

\section{Conclusions}

\label{sec:conclusions}
 
Most high-accuracy approaches to LLM-assisted mathematical modelling rely on specialisation, such as dedicated fine-tuning, expert-curated knowledge, a single paradigm or solver, or problem-specific architectures, and are validated mainly on academic benchmarks. Meanwhile, agentic AI in manufacturing has largely operated around pre-existing optimisation models rather than building them. This work has explored a different role for agentic AI, namely the generation of the model: whether general-purpose LLMs, used without any training and orchestrated as agents, can formulate and implement CP models for industrial scheduling problems from nothing more than a natural-language description and the schema of the instance database. CP was chosen deliberately, since it is a paradigm of proven industrial strength in which LLM-assisted modelling remains far less explored than in MILP.
 
To this end, two agentic workflows were proposed, a multi-agent pipeline that separates formulation from coding and a single agent that performs both tasks, and compared with a direct LLM baseline. The formulation was entrusted entirely to the reasoning capability of the LLMs, without any formulation hint, whereas the implementation, where hallucinated and misused API calls had been observed most often, was supported by an MCP server that gives the agents on-demand access to the \texttt{docplex.cp} documentation through discovery and precision tools and a retrieval protocol enforced within the server. Because the corpus, its chunking and its retrieval are built automatically from the reference documentation, without embeddings or an expert-curated taxonomy, and because each stage of the workflow is an independent module, the design can be transferred to other solvers and optimisation paradigms by parsing their documentation, and its catalogue of solved examples can be extended with models previously developed and validated by the user.
 
The evaluation on six industry-oriented problems, three LLMs and 162 single-attempt generations shows, first, that formulation is largely within reach of current LLMs. On the flow shop with sequence-dependent changeovers and no-wait coupling, the weighted-tardiness job shop, the flexible job shop with machine eligibility and a cumulative operator pool, and the basic AGV transport and picking problem (PC1--PC4), 97.2\% of the generated models were correctly formulated. Moreover, all three LLMs formulated every assessable run correctly, so the choice among the models evaluated makes little difference at this level. The limit lies in the two extended intralogistics problems, where optional transport intervals coexist with integer allocation quantities, binary assignments, presence links and more than 2\,000 constraints: formulation accuracy falls to 44.4\% in PC5 and 14.8\% in PC6, and the remaining models are over-constrained or structurally wrong. Formulation accuracy also barely depends on the workflow (77.8\% in both agentic workflows against 68.5\% for the direct LLM), which confirms that it is governed by the reasoning capability of the LLM.
 
The agentic approach makes its difference in the implementation. The share of scripts that run correctly exactly as generated rises from 14.8\% with the direct LLM to 33.3\% with the single agent and 59.3\% with the multi-agent workflow, a fourfold improvement that reaches 80.6\% on PC1--PC4, while the share of runs lost only to coding errors falls from 53.7 to 18.5 percentage points. The gain holds for every LLM, and the multi-agent workflow is the only configuration in which all three replicates of a setup produced a correct script, which indicates that separating formulation from coding is beneficial in itself. This gain has a cost: the multi-agent workflow requires on average 370\,s and 0.73 million tokens per run, against 142\,s and about 12\,000 tokens for the direct LLM, because the retrieved documentation is re-sent at every turn of the tool-calling loop. The cost is determined by the LLM far more than by the problem: DeepSeek V4 Pro consumes about seven times more tokens than Gemini 2.5 Pro, owing to its three to four times more frequent tool calls, whereas Gemini 2.5 Pro is, in the multi-agent workflow, both the cheapest model (0.18 million tokens per run) and the most accurate one (66.7\%).

Considering that the output is an optimisation model that would otherwise require substantial expert effort, a generation time of a few minutes and a cost of roughly US\$0.3--1 per model at current list prices remain highly competitive.
 
Finally, the error analysis shows that implementation failures are loud, concentrated and recoverable. Of the 104 incorrect scripts, 89 stop with an explicit error or an unusable output rather than a silently wrong schedule; the 159 distinct coding errors found in 94 runs amount to fewer than two per script, so that at least two thirds of these scripts contain only one or two errors, and five recurrent mistakes explain more than half of the total; and a minimal literal patch turns 63 of the 104 incorrect scripts into correct ones, raising the number of correct models from 58 to 121. Only one error in four is a hallucination in the strict sense, whereas most are misuses of existing API elements, and the most frequent single error concerns the accessors of the solve result, which lie outside the documented corpus. 

Taken together, the results show that general-purpose LLMs, without any task-specific training, orchestrated as agents and grounded in solver documentation, can already formulate classical and moderately extended scheduling problems reliably and turn a large share of them into executable CP models at the first attempt, at a moderate computational cost. Although tightly coupled intralogistics models remain an open challenge, the combination of this performance with a modular design that is not tied to a specific problem class, solver or training procedure suggests that agentic AI emerges as a promising and versatile route towards making mathematical optimisation accessible to industrial practitioners.

\section{Limitations and future directions}

Despite these encouraging results, the study also reveals limitations that point to several directions for future research. First, formulation accuracy drops sharply when assignment logic, heterogeneous variable types and thousands of constraints coexist, as in the extended intralogistics problems PC5 and PC6 (44.4\% and 14.8\%). For problems of this complexity, expert knowledge and specialised LLM-based approaches therefore remain necessary, and their combination with agentic workflows is a natural next step, for instance through expert review of the explicit formulation that the multi-agent pipeline produces before any code is written. Second, although the agentic configurations substantially improve the implementation, even the multi-agent workflow still produces faulty code in a considerable share of runs. Since most of these failures are explicit and are resolved by minimal corrections, iterative repair loops driven by execution feedback, deliberately excluded from this single-attempt evaluation, are expected to add considerable value. Finally, a practitioner has no reference objective against which to judge a generated model, so reliable mechanisms for validating its output, such as solver-independent constraint checkers and consistency tests on the resulting schedules, will be essential for the safe industrial deployment of these approaches.

\section*{Disclosure statement}
No potential conflict of interest was reported by the author(s).

\section*{Funding}
The authors received no specific funding for this project.

\section*{Data availability statement}
The problem statements, instance databases, system prompts and generated scripts that support the findings of this study are openly available in the project repository at \url{https://github.com/angelsnfn/AgenticPScheduling} \citep{AgenticPScheduling2026}.

\section*{Declaration of generative AI use}
During the preparation of this work the authors used Claude (Anthropic) to assist with language editing of the manuscript, with the generation of LaTeX tables and figures. The large language models evaluated in the study (Gemini 2.5 Pro, GPT-5 and DeepSeek V4 Pro) are the object of the experiments and are described in Section~4. The authors reviewed and edited all content and take full responsibility for the final manuscript.

\bibliographystyle{apalike}
\bibliography{references}
\bigskip

\end{document}